\documentclass[fleqn,10pt]{wlscirep}
\usepackage[utf8]{inputenc}
\usepackage[T1]{fontenc}
\usepackage{wrapfig}
\usepackage{caption}
\usepackage{subcaption}
\usepackage{float}
\usepackage{textcomp}

\usepackage{appendix}

\title{Challenges in explaining deep learning models for data with biological variation}

\author[1,*]{Lenka Tětková}
\author[2]{Erik Schou Dreier}
\author[2]{Robin Malm}
\author[1]{Lars Kai Hansen}

\affil[1]{Department of Applied Mathematics and Computer Science, Technical University of Denmark, Kgs. Lyngby, 2800, Denmark}

\affil[2]{FOSS Analytical A/S, Hillerød, 3400, Denmark}
            
\affil[*]{lenhy@dtu.dk}


\begin{abstract}
Much machine learning research progress is based on developing models and evaluating them on a benchmark dataset (e.g., ImageNet for images). However, applying such benchmark-successful methods to real-world data often does not work as expected. This is particularly the case for biological data where we expect variability at multiple time and spatial scales. Typical benchmark data has simple, dominant semantics, such as a number, an object type, or a word. In contrast,  biological samples often have multiple semantic components leading to complex and entangled signals. Complexity is added if the signal of interest is related to atypical states, e.g., disease,  and if there is limited data available for learning.

In this work, we focus on image classification of real-world biological data that are, indeed,  different from standard images. We are using grain data and the goal is to detect diseases and damages, for example, "pink fusarium" and "skinned".
Pink fusarium, skinned grains, and other diseases and damages are key factors in setting the price of grains or excluding dangerous grains from food production.
Apart from challenges stemming from differences of the data from the standard toy datasets, we also present challenges that need to be overcome when explaining deep learning models. For example, explainability methods have many hyperparameters that can give different results, and the ones published in the papers do not work on dissimilar images. Other challenges are more general: problems with visualization of the explanations and their comparison since the magnitudes of their values differ from method to method. An open fundamental question also is: How to evaluate explanations? It is a non-trivial task because the "ground truth" is usually missing or ill-defined. Also, human annotators may create what they think is an explanation of the task at hand, yet the machine learning model might solve it in a different and perhaps counter-intuitive way.
We discuss several of these challenges and evaluate various post-hoc explainability methods on grain data. We focus on robustness, quality of explanations, and similarity to particular "ground truth" annotations made by experts. The goal is to find the methods that overall perform well and could be used in this challenging task. We hope that the proposed pipeline would be used as a framework for evaluating explainability methods in specific use cases.
\end{abstract}
\begin{document}

\flushbottom
\maketitle
\thispagestyle{empty}

\section{Introduction}
\label{sec:introduction}
While much progress in deep learning has been based on standardized benchmark data, such as MNIST \cite{deng2012mnist}, ImageNet \cite{imagenet_cvpr09} and LibriSpeech \cite{panayotov2015librispeech}, real-world applications often pose new and unexpected challenges. This is particularly the case for biological data for which we find variability at multiple time and spatial scales \cite{chen2022recent}. Benchmark data often has simple, dominant semantics, such as a number, an object type or a word, however, biological samples often have multiple semantic components leading to complex and entangled signals. Complexity is added if the signal of interest is related to abnormal states, e.g., disease,  and if there is limited data available for learning.
These important challenges are further amplified when working with the explainability of such models, as there is a large number of methods available and only limited guidance on which method can be applied under given conditions \cite{Hansen_Nielsen_Strother_Lange_2001}. In order to make decisions about which explainability method to use, it is important to develop quantitative performance metrics, an issue that has not yet been solved in benchmark nor in real-world data.

To address these important challenges, we turn to the complex biological decision problem of grain defect detection based on visual samples. This problem  shows many of the above complications.

We train networks to make decisions and propose workflows to evaluate the performance of a wide range of heatmap explainability 
methods. We identify several decisions that need to be made and obstacles on the way. We find that there is no fixed ranking of the methods for all metrics at the same time because the performance of the methods varies depending on the metric chosen. We propose a method for aggregating these results into one robust ranking specific for the data in hand. Overall, we propose a workflow for choosing explainability methods that perform well on a given dataset. However, it is quite flexible and may need further adaptation for the specific needs of the end-user.

We apply a method for aggregating heatmaps from different explainability methods into a mean heatmap \cite{Hansen_Nielsen_Strother_Lange_2001}. We find that for many of the metrics, the mean map performs better than a randomly chosen method. Hence, it might be used as a relatively 'safe' strategy when there is little time for evaluating the performance of the explainability methods.

We identify additional open questions and suggest some answers. For example: How should one define a ground truth for explainability? What explainability method should one use? How does the choice of hyperparameters influence the explanation? Can the way we plot an explanation affect the message delivered to an end-user? 
\begin{figure}[htbp]
    \includegraphics[width=\linewidth]{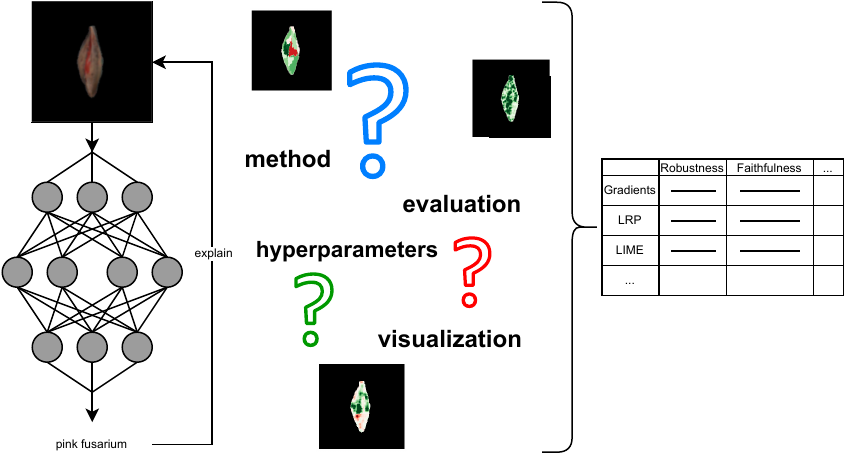}
    \caption{Our paper explores challenges that one faces when explaining image classifiers, especially for the application on data with biological variation. We train a convolutional network for grain defect detection. We discuss some of the challenges and questions that arise in connection to applying post-hoc explanation methods on this model. For example, the choice of the explainability method is crucial.  We evaluate the quality of explanations, perform an extensive analysis of some of these choices, and present the results, showing how big an impact each choice has.}
    \label{fig:Figure1}
\end{figure}

\autoref{fig:Figure1} depicts the content of this paper. It is structured in the following way: in \autoref{sec:challenges}, we identify and discuss some of the challenges of using post-hoc explainability methods. In \autoref{sec:methods}, we describe all the methods used in this paper, among others the evaluation tools. \autoref{sec:experiments} shows the overview of the results and the detailed results are deferred to Appendix. We conclude with the discussion in \autoref{sec:discussion}.

\section{Challenges}\label{sec:challenges}
This section presents a subset of all the challenges one has to face when using post-hoc explainability methods. First, we discuss how to evaluate the quality of an explanation and how complex it might be to define a relevant ground truth. A core challenge are the hyperparameters -- where do we have to specify them and how much does this choice influence the result? Next, we discuss problems regarding visualization of an explanation: channel pooling and normalization of the values. Finally, we show that explanations from two or more explainability methods cannot be directly combined and propose a method for aggregating multiple explanations.

\subsection{Challenges of evaluation}
\begin{figure}[htbp]
    \centering
    \begin{subfigure}{.45\linewidth}
    \includegraphics[width=\linewidth]{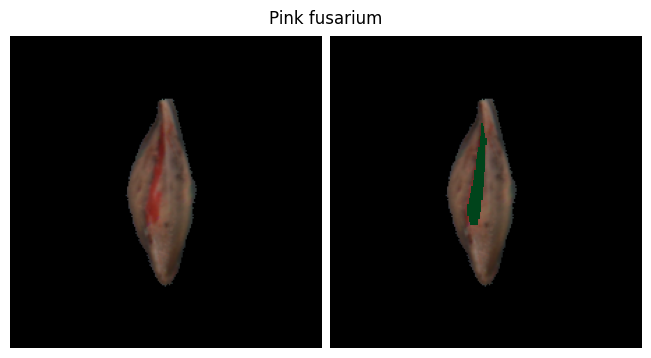}
    \end{subfigure}
    \begin{subfigure}{.45\linewidth}
    \includegraphics[width=\linewidth]{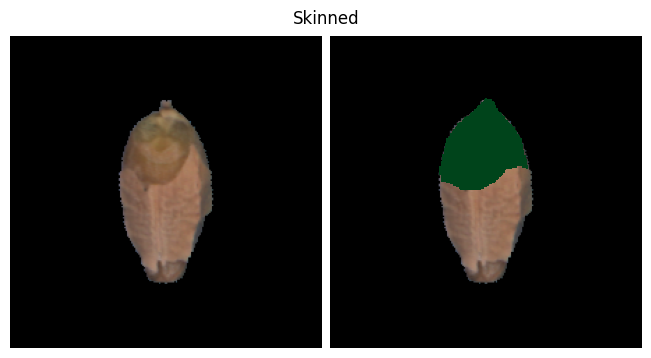}
    \end{subfigure}
    \caption{Examples of images of grains with pink fusarium (left) and skinned (right) with human annotations.}
    \label{fig:annotathtions}
\end{figure}

Unlike most machine learning tasks, the field of explainable AI suffers from a lack of ground truth. While labeling ground truth for classification involves a simple decision task, ground truth labeling for explanations requires a more detailed causal model of the imagined object. Moreover, deep learning models can potentially learn to solve a given task based on features and patterns that are very different from human cognition. Hence, an explainability method might explain the inner processes of the models well, but we humans might discard it as non-sensical. Therefore, it is not easy to assess whether a given explanation is good or not. Or, given two explanations, which one is better? This assessment is very subjective. There are possible partial solutions:
\begin{enumerate}
    \item Evaluate the explanations without knowing what the correct one looks like. There are many metrics for quality evaluation, e.g., sensitivity \cite{bhatt2020evaluating}, complexity \cite{yeh2019fidelity}, pixel-flipping \cite{Bach_Binder_Montavon_Klauschen_Muller_Samek_2015}, IROF (Iterative Removal Of Features)\cite{rieger2020irof}. Each of them measures a certain property that the explanation should fulfill (usually originating in theoretical desiderata, e.g. from \cite{Swartout_Moore_1993}). However, some of these properties are independent and an explanation scoring very well in one of them might score low in another. There is no clear consensus about which of these properties are the most important. Moreover, the evaluation metrics themselves often also require a certain design or parameter choice. \item Create ground truth explanations and compare them to the explanations generated by the explainability methods. Nevertheless, this is not easy (see the next challenge). 
    \autoref{fig:annotathtions} shows a kernel with pink fusarium and a human annotation. Note that this "ground truth" is very different from what a post-hoc explanation usually looks like. This is partially caused by a dissimilar perception of humans and machine-learning algorithms that look more at the pixel level. 
    See more details in \autoref{sec:challenges:ground_truth}.
\end{enumerate}

\subsection{Ground truth}\label{sec:challenges:ground_truth}
Getting ground truth annotations is time-consuming and expensive since it requires extensive work from humans with specialized knowledge. However, that might not be the only obstacle: even defining what the ground truth should look like might be a problem.
When grain experts assess the quality of grains, they usually look at the grain as a whole, sometimes even considering it together with other grains from the same pile. Therefore, it might be hard to pinpoint what part of the grain is the most important for the prediction.

For some diseases, the overall impression can be "the whole grain looks unhealthy". Should then the explanation be the whole grain? For example, a grain afflicted by pink fusarium looks more pinkish than a healthy grain and it is often the case for the whole kernel. Some of the damages are more localized, but even that does not ease the task. For instance, when annotating skinned grain, damage defined as "kernels from which one-third or more of the husk has been removed, or which have the husk loosened or removed over the germ" \cite{ebc_analytica, gta}, should we highlight the part of the grain where the skin is missing or the border? Moreover, to spot abnormal parts, we need to also see what the "normal" looks like because of the biological variability (Is this only miscoloring or a real morphological change?). Hence, should the annotation include only the affected part of the grain or also the fine parts? Finally, annotating is very subjective because humans perceive the world differently, so annotations are likely to differ from one annotator to another.

Other interesting approaches that would avoid the complexity of ground-truth annotations are concept-based explainability methods (e.g., \cite{kim2018tcav}). For example, if we could represent the concept of pink fusarium just by showing example images, we might explore the global functioning of the model concerning this concept and gain insights into the whole process without annotating each individual image.

\subsection{Hyperparameters} \label{sec:challenges:hyperparameters}
\begin{wrapfigure}{R}{0.5\textwidth}
    \centering
    \includegraphics[width=0.48\textwidth]{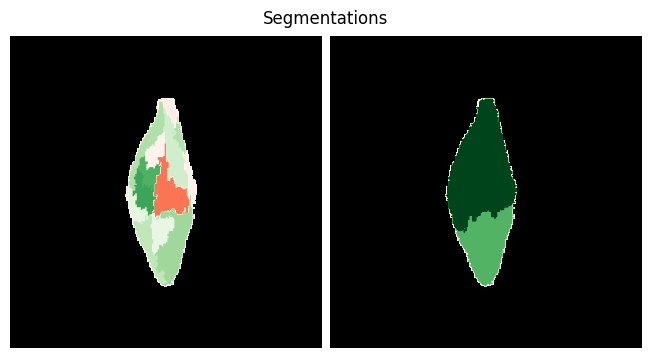}
    \caption{LIME explanations of the same image with two different segmentations.}
    \label{fig:LIME}
\end{wrapfigure}

\begin{figure}[htbp]
    \centering
    \includegraphics[width=0.8\linewidth]{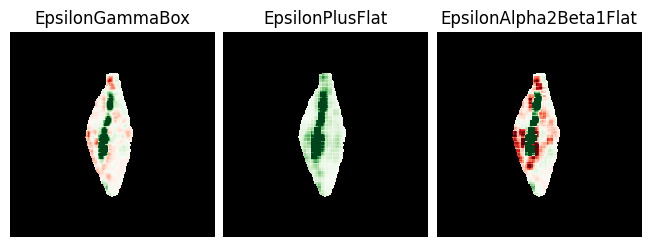}
    \caption{Heatmaps of three LRP methods used in this paper. These methods differ only in the propagation rules used for each type of layer. For details about the individual methods see \autoref{sec:expl_methods}.}
    \label{fig:LRP}
\end{figure}
Many of the post-hoc explainability methods have lots of hyperparameters to choose from. The result (heatmap) is highly dependent on these hyperparameters. 
For example, methods like LIME \cite{lime} and SHAP \cite{shap} depend on the segmentation of the image because they assign one value for the whole segment. \autoref{fig:LIME} shows LIME explanations of the same grain and model but with different image segmentation. Other methods (e.g., Integrated Gradients  \cite{sundararajan2017axiomatic}) require a baseline (as a starting point from which integral is computed), sliding window shapes (Occlusion \cite{zeiler2014visualizing}) or a decision whether to multiply the result by inputs or not (e.g., Integrated Gradients, DeepLIFT \cite{shrikumar2017learning}). In Layer-wise Relevance Propagation (LRP) \cite{bach2015pixel, kohlbrenner2020towards}, one has to choose a rule for each layer. The results depend on this choice, as illustrated by \autoref{fig:LRP}.

In general, the more complicated the method is, the more hyperparameters it has. The more complicated methods usually give better results, but at the cost of finding a suitable combination of hyperparameters. 
Since there is no single metric measuring how good overall the explanation is, automatic hyperparameter tuning is not possible for explanations. The development of such a universal metric would facilitate further advancement in this field.

Moreover, quality evaluation metrics also depend on the choice of hyperparameters. We take pixel-flipping and IROF as an example. These methods gradually replace some pixels with a certain value and measure how much these modifications change the output probability of the target class. Many questions arise from this vague description: What should be the value (color) that we use for flipping the pixels? Should we first flip the most important or the least important pixels? If we plot the changed probabilities as a curve, how do we get a score for further comparisons? If we flip whole segments instead of pixels (in IROF), what should be on the x-axis of the plotted curve (number of segments flipped, proportion of the image, proportion of the values of the explanations...)?

In conclusion, there are an immense amount of hyperparameters to tune both for the explainability methods and the evaluation metrics. Many of them drastically change the result, so one has to be careful when choosing them and comparing one to another.

\subsection{Channel pooling}
\begin{wrapfigure}{R}{0.5\textwidth}
    \centering
    \includegraphics[width=0.48\textwidth]{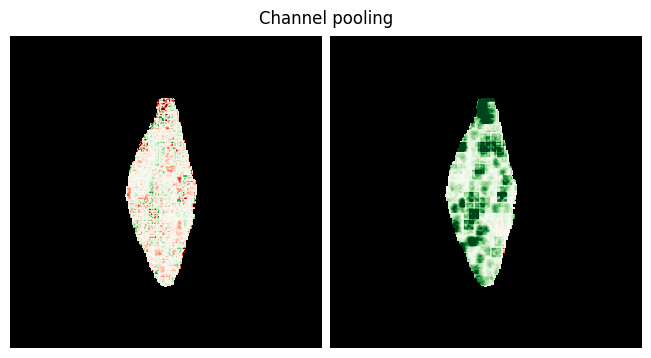}
    \caption{Mean (left) and max (right) pooling of Gradients explanation.}
    \label{fig:channels}
\end{wrapfigure}
Many of the explainability methods give an explanation that has the same shape as the original image, i.e., three channels. However, we can visualize only one channel at a time. Therefore, we need to either plot every channel separately or pool the channels to get just one.
Inspired by the section on relevance pooling in \cite{arras2020ground}, we explore the following pooling methods: \textit{mean pooling}: $R_{\text{mean}} = \frac{1}{C} \sum_{i=1}^C R_i$; \textit{max pooling}: $R_{\text{max}} = \max_i R_i$; \textit{max abs pooling}: $R_{\text{max abs}} = \max_i \vert R_i \vert$ and \textit{$\ell_2$-norm pooling}: $R_{\ell_2} = \sqrt{\frac{1}{C} \sum_{i=1}^C R_i^2}$.

\autoref{fig:channels} shows an explanation of the same grain with mean and max pooling.
One has to be careful because the channel pooling might change the meaning of the explanation: some methods show positive and negative significance but, for example, max abs and $\ell_2$-norm pooling output only positive values, therefore disregarding the sign of the contribution and focusing only on the magnitude.
\begin{figure}[htbp]
    \centering
    \includegraphics[width=\linewidth]{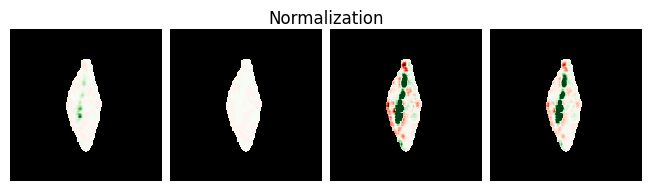}
    \caption{LRP - EpsilonGammaBox explanation with mean channel pooling. The explanation is the same in all four cases, the difference is only in normalization. Normalizing factors are from left to right: maximum over the explanation, maximum over a set of explanations, the 99th percentile of the explanation, and 99th percentile over the set of explanations of this method.}
    \label{fig:normalization}
\end{figure}
\subsection{Visualization}
A problem less investigated in  current research focus is how to visualize the explanations (that is, match the heatmap values to gray scale for plotting). The magnitude of the explanations reflects the importance of each pixel on the prediction. Within a given explainability method, these magnitudes should be comparable across images. However, when visualized separately, the maximum values of each explanation get the same color shade, and, therefore, seem to be equally important even though the magnitudes may differ drastically. Our solution is to fix the normalizing value for a set of explanations (e.g., by computing the overall maximum of all explanations in the set). In this way, the explanations are also visually comparable. However, to reduce the effects of outliers, we use the more robust value of the 99'th percentile. \autoref{fig:normalization} shows different normalization strategies. In all figures in this paper, we use a color map from red (-1), over white (zero) to green (1). The darker the color is, the higher the absolute value it represents.
\begin{wrapfigure}{R}{0.5\textwidth}
    \centering
    \includegraphics[width=0.48\textwidth]{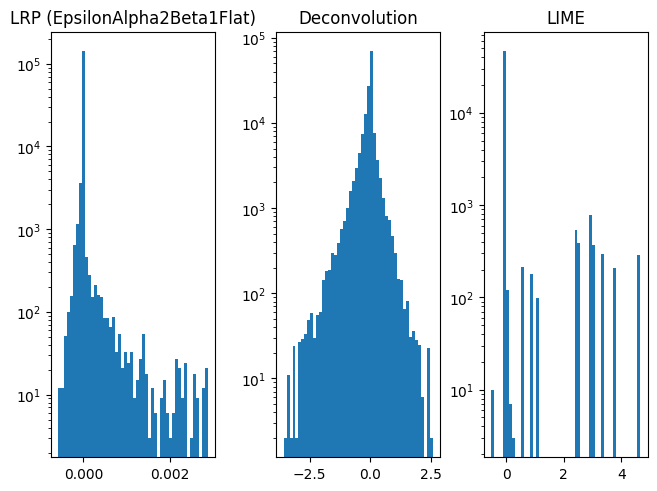}
    \caption{Histograms of per-pixel values of each explainability method for one image.}
    \label{fig:histograms}
\end{wrapfigure}
\subsection{Aggregating multiple explanations}\label{sec:challenges:mean}
Each explainability method gives an output on different scales (see \autoref{fig:histograms}). This poses a problem not only on visualization as we discussed in the previous point, but also if we want to aggregate multiple explanations.
Ensemble methods have been used to improve the performance \cite{sagi2018ensemble} because averaging can never produce a summary that is worse than the average of the participating models. Ensembling has also been used in the field of explainable AI. For instance, one can average explanations produced by one method and applied on many different versions of the same model (e.g., \cite{Ahamed_Alipour_Kumar_Soltani_Pazzani_2022, woerl2023initialization}). There is no problem with scaling in this case if we expect one method to be consistent over various inputs.
In this paper, we average over different explainability methods. Rieger and Hansen in \cite{rieger2019aggregating} do that as well but they simply compute an average without balancing the scales first. They find that the aggregation of explanations is more robust and well-aligned with human explanations.

Inspired by histogram equalization (Section 2 in \cite{Hansen_Nielsen_Strother_Lange_2001}), we propose the following: we transform all explanations into the same distribution (in our case, the normal distribution with mean 0 and standard deviation of 1). This procedure keeps the ordering of the elements the same while transforming them to the same scale. We then take a per-pixel average over all the explainability methods.
Other possibilities for aggregating exist: one might use voting instead of the mean or consider also aggregating explanations of larger segments (e.g. those produced by LIME and SHAP) instead of per-pixel values.

\section{Methods}\label{sec:methods}
\subsection{Data}
The grain image data used in this paper was obtained from the FOSS's EyeFoss™ image database. EyeFoss is an instrument for objective grain quality estimation using image-based classification of grain types and grain damages \cite{EyeFoss}. From the EyeFoss database, two well-known and well-described barley defects important for the malting process, were selected: pink fusarium infection and skinned barley. Pink fusarium (fusarium head blight, or just blighted kernels)  is a severe fungi infection in grain that causes significant damage to the crops and potentially produces vomitoxin, i.e., deoxynivalenol (DON) \cite{kumar2007identification} that affect the health of humans and animals \cite{fermentation4010003}. The presence of pink fusarium will result in the downgrading of grains resulting in lower prices or even loads being discarded. Fusarium can affect the appearance of barley in multiple ways, but the most significant is the potential discoloring of the barley kernel with clear pink areas \cite{mcmullen2008fusarium,gta}. In this work, we used the definition that pink fusarium is present if kernels have pink markings.

Skinned kernels are barley kernels where part of the skin has been lost. Skinning of kernels affects the malting properties of the barley kernels and, if a large fraction of a barley load is skinned, it may cause farmers to receive a lower payment. In this work, we have used the following definition of skinned kernels: a) 1/3 or more of the total surface area of the husk is missing. b) Germ Exposed - The husk is removed from the germ end of the grain or has been damaged other than Shot or Sprouted or the germ itself has been removed. c) Kernels may or may not be dark under the husk.

The dataset we used in this paper is a private dataset of images of grains. Each grain contains information about what type of grain it is (e.g., barley or wheat). It might also contain a positive or negative label about grain damages. All the labels were created by human experts on grains. 
The images were collected using the EyeFoss™ machine. It lets a heap of grains pass under a laser line profile scanner and multispectral camera and light setup to capture data. Each kernel is then segmented, rotated, and centered. The camera captures seven channels (UV, Blue, Green, Orange, Red, NIR and
Heightmap) but we are using only red, green, and blue channels. We mask the background of the grain.

We chose two damages (pink fusarium and skinned). We considered also the problem of classification of the type of grain, but that depends a lot on the shape and could potentially be solved with high accuracy with the masks only. Therefore, we chose damages where the color and structure of the grain are important and the shape is not.

For each damage, the dataset was assembled in the following way: we took all grains with positive labels for this damage and sampled the same amount of grains with negative labels. We then split the data into train, validation, and test sets in proportion $60:20:20$.
The numbers of images for both the "pink fusarium" and the "skinned" cases are in \autoref{tab:data}. Data used for the evaluation of the explanations are a subset of test sets.
\begin{table}[bp]
  \centering
  \begin{tabular}{l|l|l}
                    & Pink fusarium     & Skinned   \\ \toprule
    Train set       & 18828             & 10308     \\ \hline 
    Validation set  & 6276              & 3437      \\ \hline 
    Test set        & 6276              & 3427      \\ \bottomrule
  \end{tabular}
  \caption{Number of images in each dataset. The positive and negative samples are balanced.}
  \label{tab:data}
  % \vspace{-5mm}
\end{table}
For the evaluation of explanations, we sampled 400 images separately for the pink fusarium and skinned categories, out of which 200 have a positive label and 200 have a negative one. For comparing to the ground truth, we used 173 manually annotated images regarding pink fusarium and 145 annotated images for skinned damage. The annotations were collected manually by one annotator with the task assignment: "Highlight part(s) of the grain that indicate the presence/absence of the defect." Only the images where such annotation was obvious for the annotator were chosen because the rest would introduce complications connected to the definition of ground truth as discussed in \autoref{sec:challenges:ground_truth}.

\subsection{Models}
\begin{wrapfigure}{R}{0.3\textwidth}
    \centering
    \includegraphics[width=0.28\textwidth]{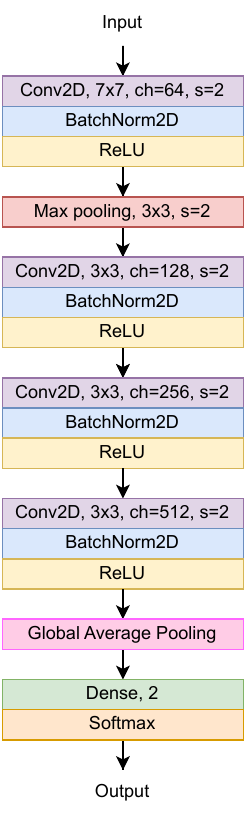}
    \caption{Model architecture used in this paper.}
    \label{fig:architecture}
\end{wrapfigure}

We use an architecture inspired by a similar architecture originally proposed for hyperspectral images in \cite{engstrom2023improving} (SimpleNet). \autoref{fig:architecture} illustrates the architecture used in this paper. The model was chosen as it can achieve good classification performance with a small training set and without transfer learning. We train one model for each disease/damage (i.e., it is a binary classification).
We find the best parameters for training by hyperparameter search. \autoref{tab:hyperparameters} in \autoref{sec:training_hyperparameters} 
shows the chosen hyperparameters.
The test-set accuracy of the model trained on "pink fusarium" data is $96.23\%$ and it is $95.78\%$ in the case of the model trained on "skinned" data.

\subsection{Explainability methods} \label{sec:expl_methods}
We investigated the following post-hoc explainability methods: Gradients \cite{simonyan2013deep}, Input x Gradients \cite{simonyan2013deep}, Integrated Gradients \cite{sundararajan2017axiomatic}, Guided Backpropagation \cite{springenberg2014striving}, Deconvolution \cite{zeiler2014visualizing}, DeepLIFT \cite{shrikumar2017learning}, Guided Grad-CAM \cite{selvaraju2017grad}, Occlusion \cite{zeiler2014visualizing}, LIME \cite{lime}, SHAP \cite{shap} and three variants of Layer-wise Relevance Propagation (LRP)\cite{bach2015pixel, kohlbrenner2020towards} composites:
EpsilonPlusFlat (LRP-$\varepsilon$-rule for dense layers, LRP-$\alpha, \beta$ ($\alpha=1, \beta=0$), also called ZPlus rule, for convolutional layers, and the flat rule for the first linear layer), 
EpsilonGammaBox (LRP-$\varepsilon$-rule for dense layers, the LRP-$\gamma$-rule ($\gamma=0.25)$ for
convolutional layers, and the LRP-$Z^B$-rule (or box-rule) for the first layer) and 
EpsilonAlpha2Beta1Flat (LRP-$\varepsilon$-rule for dense layers, LRP-$\alpha, \beta$ ($\alpha=2, \beta=1$) for convolutional layers and the flat rule for the first linear layer) \cite{Montavon_Binder_Lapuschkin_Samek_Muller_2019}.
Moreover, one has to make sure that the model architecture is compatible with the rules used in LRP. One of the problematic layers is batch normalization \cite{ioffe2015batch}. We use the SequentialMergeBatchNorm canonizer to merge the batch normalization with the preceding convolutions.

We used Zennit \cite{anders2021software} to generate LRP explanations, official repositories for SHAP and LIME, and Captum \cite{kokhlikyan2020captum} for the rest of the explainability methods.

The hyperparameters discussed in \ref{sec:challenges:hyperparameters} were chosen as follows. We set the baseline for Integrated Gradients to be 0, the sliding window shape for Occlusion $(3, 3, 3)$, and the size of the neighborhood to learn the linear model in LIME to 1000. We multiply both Integrated Gradients and DeepLIFT with input values.
We tried many segmentation methods and combinations of hyperparameters and chose one that consistently segments the kernel into several smaller segments. The chosen method is quickshift \cite{vedaldi2008quick} clustering based on color-space proximity with kernel size 25 and max distance 10. Finally, we set the whole background to be a single segment.

\subsection{Methods for evaluating the quality of explanations}\label{sec:methods_quality}
We evaluate the explanations on several desiderata: robustness to data augmentation, faithfulness, complexity, robustness, and similarity to the ground truth annotations.

\paragraph{Quality evaluation}
First, evaluate the robustness of explanations to various data augmentations. We replicate the experiments from \cite{tetkova2023robustness} and compare the results for a toy dataset (ImageNet) from \cite{tetkova2023robustness} and a real-world one (grains). We choose six augmentation methods: changes in brightness, hue and saturation, further rotation, translation, and scale.
For a given image, we create its augmented versions, generate explanations of these augmented images, and compare them to the explanation of the original image. The six chosen data augmentation techniques can be divided into two groups: invariant (the explanation of the augmented image should be the same as the explanation of the original one) and equivariant (the explanation of the augmented image should be the same as the augmented explanation of the original image). We consider the change of brightness, hue, and saturation as invariant methods and rotation, translation, and scaling as equivariant ones.

We consider only correctly classified images because we are more interested in the cases where the model "knows" the correct answer. Moreover, for misclassified cases, the probability of the target class might increase after alternating the image, leading to problems with score definition. For each data augmentation, we choose an interval of the augmentation parameter such that the probability of the target class drops on average by 0.1. We divide the interval into equidistant segments. We then plot the Pearson correlation coefficients between the (augmented) explanation of the original image and the explanation of the image augmented by each of the chosen values (e.g., rotated by $10$ degrees). We create a curve by linear interpolation and shift it in such a way, that the highest value is 1 (since we are interested in the shape of the curve and not the specific values). Next, we compute the area under the curve. To ensure independence on the predicted probabilities, we divide this number by area under the curve of similarly plotted normalized probabilities. The result is usually a number between $0$ and $1$ with $1$ meaning that the explanations are as robust as the probabilities of the classifier. Values higher than $1$ mean that the explanations are more robust than the predictions themselves.
More details can be found in \cite{tetkova2023robustness} and the corresponding repository \cite{Tetkova_Robustness_of_explanations_2023}. 

Next, we evaluate robustness in general (i.e., stability to slight perturbations of the input) using the Sensitivity \cite{yeh2019fidelity} score that measures the average sensitivity of an explanation using a Monte Carlo sampling-based approximation.

One of the desiderata is faithfulness, i.e., how much the explanation follows the behavior of the model. We choose two metrics coming under the faithfulness category: pixel-flipping\cite{Bach_Binder_Montavon_Klauschen_Muller_Samek_2015} and IROF\cite{rieger2020irof}.
We compute pixel-flipping in the same way as Tětková and Hansen did in \cite{tetkova2023robustness}. First, we add negligible random values to the explanation values of all pixels to break ties (for example, in the case of LIME and SHAP). This might influence the results, especially if the segments are large. We flip the most important pixels first and replace them with the mean value of the image. For each perturbed image, we normalize its probability of the target class by the probability of the original image of the target class. We plot these values as a curve by linear interpolation (with the percentage of pixels flipped on the x-axis). The pixel-flipping score for one image is defined as the normalized area over the curve (up to $1$) from zero to the first $20\%$ pixels flipped. We average these scores across all correctly classified images.
 
Some explainability methods produce pixel-level explanations, whereas others operate on segments. Therefore, it makes sense to flip whole segments at once instead. We fix the segments for each image. For the pixel-level explainability methods, we compute the mean value of the attributions in each segment. We iteratively replace the whole segments from most to least relevant by the mean value of the image. As with pixel flipping, we plot the normalized probabilities as a curve by linear interpolation. However, the x-axis shows the number of segments flipped. The score is then the area over the curve for the whole curve (averaged across all correctly classified images).

Another desiderata for explanations might be complexity: only a few features should be used for the explanation. We evaluate this using a metric called complexity \cite{bhatt2020evaluating} which calculates the entropy based on the individual contributions of each feature to the overall magnitude of the attribution.

\paragraph{Similarity to ground truth}
We also evaluate how well the explanations align with the ground truth annotations. First, we compute the area under the Receiver Operating Characteristic Curve (ROC-AUC)\cite{BRADLEY19971145}. If only one type of annotation (positive/negative) is present in the annotation, we treat it as a binary classification problem. If both positive and negative regions are present in the annotation, we treat it as a classification into three classes (positive, neutral, and negative) and compute the AUC as one-vs-one, i.e., averages of all possible pairwise combinations of classes\cite{hand2001simple}.
The second evaluation metric is Relevance Mass Accuracy \cite{arras2020ground} which measures the proportion of positively attributed attributions within the ground truth mask in relation to the total positive attributions.

We used the Quantus library \cite{hedstrom2023quantus} for quality evaluation. However, we changed the code for pixel-flipping and IROF to fit our specific definitions.

\paragraph{Final ranking definition}
We have in total 12 metrics for quality evaluation. Moreover, if we also take into account four different types of pooling, we get 24 metrics. Each one is measuring slightly different aspects of explanations. We propose a method for how to rank the explainability methods based on all the metrics at once. The core idea is mean reciprocal rank (MRR) used for example in information retrieval \cite{liu2009learning}. It is defined as $ MRR(m) = \frac{1}{N}\sum_{i=0}^N \frac{1}{rank_i(m)}$,
where $N$ is the number of rankings (metrics) available, $m$ is an explainability method and $rank_i(m)$ is the rank of method $m$ in the $i-$th ranking.

Here, we use the means over all input data. We use median imputation \cite{zhang2016missing} in case of missing values. We chose a single imputation for simplicity and we decided to prefer median over mean because it is more robust to outliers. We also add a small noise to all the values to randomly break ties. 

We group the metrics into three groups: robustness to data augmentation, quality of explanation (without ground truth), and compliance to ground truth. Since we would like each group to have the same vote regardless of the number of elements in each group, we first aggregate the rankings within each group. We sort the exaplainability methods by MRRs within each group. In this step, we get only three rankings (one per group). We repeat the computation of MRRs and, finally, sort the explainability methods by these MRRs.

One slight modification of this rank evaluation is to use Monte Carlo simulation \cite{monte_carlo}. Using only the means over all data, without uncertainties, might result in noisy ranking because it depends on the images we use for evaluation. Therefore, we see each score as a normal distribution (with mean being the mean over all data and standard deviation defined as the standard error of the mean over all data, i.e., the uncertainties reported in the results). For each of $n$ repetitions (we set $n=100$), we draw a sample from each distribution and repeat the whole ranking procedure with these samples instead of the means. We get $n$ rankings that we, again, aggregate into one using MRRs. This approach is more sensitive to overlapping distributions and, therefore, delivers a result that does not depend on the individual input images. We report the final rankings both with and without Monte Carlo simulation to see the difference.

\section{Experiments}\label{sec:experiments}
The detailed results for all metrics and explainability methods are in \autoref{sec:tables}. 
Here, we report the main outcomes and the aggregated rankings.

\subsection{Robustness to data augmentation} \label{sec:robustness}
When compared to the results on ImageNet from \cite{tetkova2023robustness}, we see that the results on robustness look very similar. Specifically, intervals (determined by the drop in probability) and results are close for both pink fusarium and skinned damages. This suggests that robustness to data augmentations does not heavily depend on the data distributions. Moreover, invariant methods are almost as robust as the predictions, whereas the equivariant methods lack in both cases. The best-performing methods are LRP: EpsilonPlusFlat and Deconvolution (compared to Guided Backpropagation instead of Deconvolution in the case of ImageNet).

\subsection{Evaluation of the quality (without ground truth)}
Regarding faithfulness, the scores differ notably, ranging from $~0.1$ to $~0.85$. Overall, the results for pink fusarium are higher than for skinned. This highlights the importance of testing different methods under the same conditions if we want to compare them. In IROF, LIME and SHAP perform the best. This might be caused by the fact that they produce explanations for the whole segments, whereas for other methods, we estimate the importance of each segment by averaging values for all pixels. Analogously, the pixel-wise explainability methods perform better at pixel-flipping (compared to the segment-wise methods where we randomly sort pixels within each segment).

We can observe that there is no clear winner in terms of the channel poolings. Some methods perform better with one type of pooling and others with another. Since the choice of channel pooling changes the meaning of the explanations, one should choose the pooling first based on the desired properties and then find an explainability method that performs well with this pooling type. This is also the reason why we decided to include all four types of pooling into the final aggregation of the results.

A notable observation was the superior performance of aggregated explanations (discussed in \autoref{sec:challenges:mean}) over individual methods, which, in some instances, surpassed even the best-performing single method. This trend underscores the potential benefits of ensemble approaches in explanation generation. When aggregating multiple evaluation scores, we need to be careful. If we rank the methods only using the means, we might lose important information. The uncertainty intervals are often overlapping, so the Monte-Carlo sampling would produce different rankings every time.

If we evaluate the pixel-flipping also on the ground-truth annotations, there is a big difference between the result for pink fusarium ($0.180\pm 0.016$) and skinned ($0.499 \pm 0.018$). It seems that the human annotation for skinned somehow follows the behavior of the model but not at all for pink fusarium. This might happen because the kernels in the category "skinned" are usually only mechanically damaged, but the kernels with pink fusarium may also have other diseases and generally look unhealthy (so the problem is more complicated).

\subsection{Evaluation: "ground truth"}
We compared the explanations to the human annotations using two metrics: ROC-AUC and Relevance Mass Accuracy (see \autoref{sec:tables} for the details). 
Based on the results, we see that these two metrics supplement each other, so we cannot choose only one of them without loosing valuable information.

We observe that LRP (EpsilonPlusFlat) gives the best results in all cases, meaning that its explanations agree the most with the human-created annotations. Overall, if we consider the challenges related to the definition of ground truth (\autoref{sec:challenges:ground_truth}, the results are surprisingly good - it seems that the explanations are quite aligned with the ground truth. This is an important observation for exploring human-machine alignment.

Some of the methods are robust to the type of channel pooling used (e.g., LRP: EpsilonPlusFlat), whereas others are not (e.g., Gradients, GuidedBackprop...). Overall, $l_2$-norm channel pooling gives the highest ROC-AUC results.

\subsection{Aggregation of the metrics}
As we saw in previous sections, the results are very different for various metrics, therefore, it is not straightforward to determine the best method. We aggregate all the rankings as described in \autoref{sec:methods_quality}.
\begin{table}
\begin{subtable}[thbp]{\textwidth}
\begin{center}
\begin{tabular}{l|l|l}
     & Means only               & Final MRR   \\  \hline
1. 	 & LRP (EpsilonPlusFlat) & 0.528  	 \\ \hline
2. 	 & SHAP & 0.403  	 \\ \hline
3. 	 & Deconvolution & 0.387  	 \\ \hline
4. 	 & LRP (EpsilonAlpha2Beta1Flat) & 0.344  	 \\ \hline
5. 	 & Guided Backpropagation & 0.292  	 \\ \hline
6. 	 & LIME & 0.234  	 \\ \hline
7. 	 & LRP (EpsilonGammaBox) & 0.225  	 \\ \hline
8. 	 & DeepLIFT & 0.180  	 \\ \hline
9. 	 & mean & 0.137  	 \\ \hline
10. 	 & Integrated Gradients & 0.129  	 \\ \hline
11. 	 & Guided Grad-CAM & 0.121  	 \\ \hline
12. 	 & Occlusion & 0.098  	 \\ \hline
13. 	 & Input x Gradients & 0.094  	 \\ \hline
14. 	 & Gradients & 0.080  	 \\ \hline
\end{tabular}\quad
\begin{tabular}{l|l|l}
 & Monte Carlo              & Final MRR   \\  \hline
1. 	 & LRP (EpsilonPlusFlat) & 0.820 	 $\pm$ 0.031  	 \\ \hline
2. 	 & LRP (EpsilonAlpha2Beta1Flat) & 0.485 	 $\pm$ 0.032  	 \\ \hline
3. 	 & SHAP & 0.470 	 $\pm$ 0.006  	 \\ \hline
4. 	 & Deconvolution & 0.269 	 $\pm$ 0.008  	 \\ \hline
5. 	 & Guided Backpropagation & 0.200 	 $\pm$ 0.004  	 \\ \hline
6. 	 & LRP (EpsilonGammaBox) & 0.190 	 $\pm$ 0.004  	 \\ \hline
7. 	 & LIME & 0.148 	 $\pm$ 0.003  	 \\ \hline
8. 	 & DeepLIFT & 0.123 	 $\pm$ 0.001  	 \\ \hline
9. 	 & mean & 0.113 	 $\pm$ 0.001  	 \\ \hline
10. 	 & Integrated Gradients & 0.102 	 $\pm$ 0.001  	 \\ \hline
11. 	 & Guided Grad-CAM & 0.100 	 $\pm$ 0.002  	 \\ \hline
12. 	 & Input x Gradients & 0.082 	 $\pm$ 0.000  	 \\ \hline
13. 	 & Occlusion & 0.075 	 $\pm$ 0.000  	 \\ \hline
14. 	 & Gradients & 0.075 	 $\pm$ 0.000  	 \\ \hline
\end{tabular}
\end{center}
\caption{Pink Fusarium: using means over available data (left) and Monte Carlo simulation (right).}
\end{subtable}
\begin{subtable}[t]{\textwidth}
\begin{center}
\begin{tabular}{l|l|l}
 & Means only              & Final MRR   \\  \hline
   1. 	 & LRP (EpsilonPlusFlat) & 0.722  	 \\ \hline
2. 	 & LIME & 0.403  	 \\ \hline
3. 	 & Occlusion & 0.325  	 \\ \hline
4. 	 & LRP (EpsilonAlpha2Beta1Flat) & 0.236  	 \\ \hline
5. 	 & SHAP & 0.226  	 \\ \hline
6. 	 & Deconvolution & 0.221  	 \\ \hline
7. 	 & mean & 0.183  	 \\ \hline
8. 	 & Input x Gradients & 0.175  	 \\ \hline
9. 	 & Integrated Gradients & 0.162  	 \\ \hline
10. 	 & Guided Backpropagation & 0.148  	 \\ \hline
11. 	 & Guided Grad-CAM & 0.134  	 \\ \hline
12. 	 & DeepLIFT & 0.120  	 \\ \hline
13. 	 & LRP (EpsilonGammaBox) & 0.118  	 \\ \hline
14. 	 & Gradients & 0.078  	 \\ \hline
\end{tabular}\quad
\begin{tabular}{l|l|l}
 & Monte Carlo              & Final MRR   \\  \hline
1. 	 & LRP (EpsilonPlusFlat) & 1.000 	 $\pm$ 0.000  	 \\ \hline
2. 	 & LIME & 0.439 	 $\pm$ 0.011  	 \\ \hline
3. 	 & Occlusion & 0.276 	 $\pm$ 0.009  	 \\ \hline
4. 	 & LRP (EpsilonAlpha2Beta1Flat) & 0.260 	 $\pm$ 0.007  	 \\ \hline
5. 	 & SHAP & 0.251 	 $\pm$ 0.012  	 \\ \hline
6. 	 & Deconvolution & 0.171 	 $\pm$ 0.002  	 \\ \hline
7. 	 & mean & 0.165 	 $\pm$ 0.006  	 \\ \hline
8. 	 & Integrated Gradients & 0.126 	 $\pm$ 0.002  	 \\ \hline
9. 	 & Input x Gradients & 0.115 	 $\pm$ 0.005  	 \\ \hline
10. 	 & Guided Backpropagation & 0.103 	 $\pm$ 0.001  	 \\ \hline
11. 	 & DeepLIFT & 0.095 	 $\pm$ 0.003  	 \\ \hline
12. 	 & Guided Grad-CAM & 0.089 	 $\pm$ 0.001  	 \\ \hline
13. 	 & LRP (EpsilonGammaBox) & 0.088 	 $\pm$ 0.003  	 \\ \hline
14. 	 & Gradients & 0.071 	 $\pm$ 0.000  	 \\ \hline
\end{tabular}
\end{center}
\caption{Skinned: using means over available data (left) and Monte Carlo simulation (right).}
\end{subtable}
\caption{Final ordering of explainability methods}
\label{tab:ranking_final}
\end{table}
The aggregated rankings are in \autoref{tab:ranking_final}. Rankings for skinned and pink fusarium are similar from afar: LRP (EpsilonPlusFlat) is always the best and Gradients is always the worst. However, they differ if we take a closer look and compare each pair of methods. This suggests that even two types of grain damage are differently hard to explain and highlights the necessity to perform this evaluation for each problem individually.

Different variations of LRP are spread throughout the whole ranking: for example, EpsilonPlusFlat is always the best, whereas EpsilonGammaBox is the second-to-last for the skinned case. Therefore, a good combination of hyperparameters is crucial (as discussed in \autoref{sec:challenges:hyperparameters}).

As we expected from the overlapping distributions, the results from Monte Carlo simulations are different than when simply taking the means. We could rely on the means only if the distributions were sufficiently separated. In our case, sampling from the distributions is necessary for reliable aggregation of the metrics.

\section{Discussion and conclusion}\label{sec:discussion}
Data with biological variation are harder to classify and explain than standard benchmark datasets. 
We introduced several challenges that one has to face when working with post-hoc explanations: means of evaluation, defining ground truth, hyperparameter choice, channel pooling, and visualization. As we saw from the examples, individual choices have a big impact on the explanation. Without careful documentation of all the hyperparameters, results from two separate experiments cannot be compared in general. This causes problems with evaluating new results, and, therefore, hinders the progress in the field of XAI.

Based on the results, we recommend using LRP: EpsilonPlusFlat for the grain data and our choice of model architecture. For any other data and models, the whole workflow needs to be repeated followed by a careful analysis of the results.
The presented procedure should be taken predominantly as a framework for evaluating explainability methods on non-standard data because the results are likely to be different when applied to other images with different properties. The ranking of the explainability methods also depends on the model we are explaining because the applicability of some explainability methods is closely linked to model architectures: it may work better for some layer types than others. For instance, the LRP method relies heavily on the canonization of the model and suitable rules.

We observed that the choice of the best explainability method is hard even with the evaluation metrics (some methods are better in some aspects and others in other criteria). We proposed a method for aggregating various explainability methods. This aggregated method is on average better than any randomly chosen method. We hypothesize that the relatively low position in the final ranking is due to the imputation and lack of results for all metrics. 

In this paper, we did not take into account the time it takes for each method to produce an explanation. The differences between the fastest ones (e.g., Gradients) and the slowest ones (e.g., Occlusion) are immense, so this aspect might also need to be taken into account for larger numbers of data.

The challenges presented in this paper have an impact on the quality and usability of AI applications. We hope this work can lead to more focus on the problems and help find productive solutions for real-world applications.

 \bibliography{refs}

\section*{Acknowledgements}
This work was supported by the DIREC Bridge project Deep Learning and Automation of Imaging-Based Quality of Seeds and Grains, Innovation Fund Denmark grant number 9142-00001B. This work was supported by the Pioneer Centre for AI, DNRF grant number P1 and the Novo Nordisk Foundation grant NNF22OC0076907 ”Cognitive spaces - Next generation explainability”.

\section*{Author contributions statement}
L.T. ran the experiments, wrote the main manuscript text, and prepared all the figures.\\
E.S.D. wrote the description of grain data.\\
R.M. created the ground truth annotations. \\
L.K.H. supervised the project and contributed to the manuscript. \\
All authors participated in the discussions about the content and design of experiments. All authors reviewed the manuscript.

\section*{Competing interests}
The author(s) declare no competing interests.

\section*{Data availability statement}
Data supporting this study cannot be made available due to commercial reasons.
\clearpage
\begin{appendix}
\section*{Appendix}
\section{Training hyperparameters}\label{sec:training_hyperparameters}
\begin{table}[htp]
  \centering
  \begin{tabular}{l|l|l}
                    & Pink fusarium     & Skinned   \\ \toprule
    batch size      & 64                & 64     \\ \hline 
    $\#$ of epochs     & 100               & 100      \\ \hline 
    LR scheduler    & reduce on plateau & cosine with restarts \\ \hline
    LR at start     & 0.02082           & 0.00383 \\ \hline
    weight decay    & 0.00184           & 0.78410      \\ \bottomrule
  \end{tabular}
  \caption{Training hyperparameters for both data categories.}
  \label{tab:hyperparameters}
\end{table}

\section{Figures of explanations}\label{sec:figures}
\subsection{Pink Fusarium}
\autoref{fig:PF_examples} shows examples of grains with pink fusarium disease, ground truth annotations and explanations from all methods used in this paper.
\begin{figure}[htbp]
\centering
    \begin{subfigure}[t]{0.7\linewidth}
    \includegraphics[width=\linewidth]{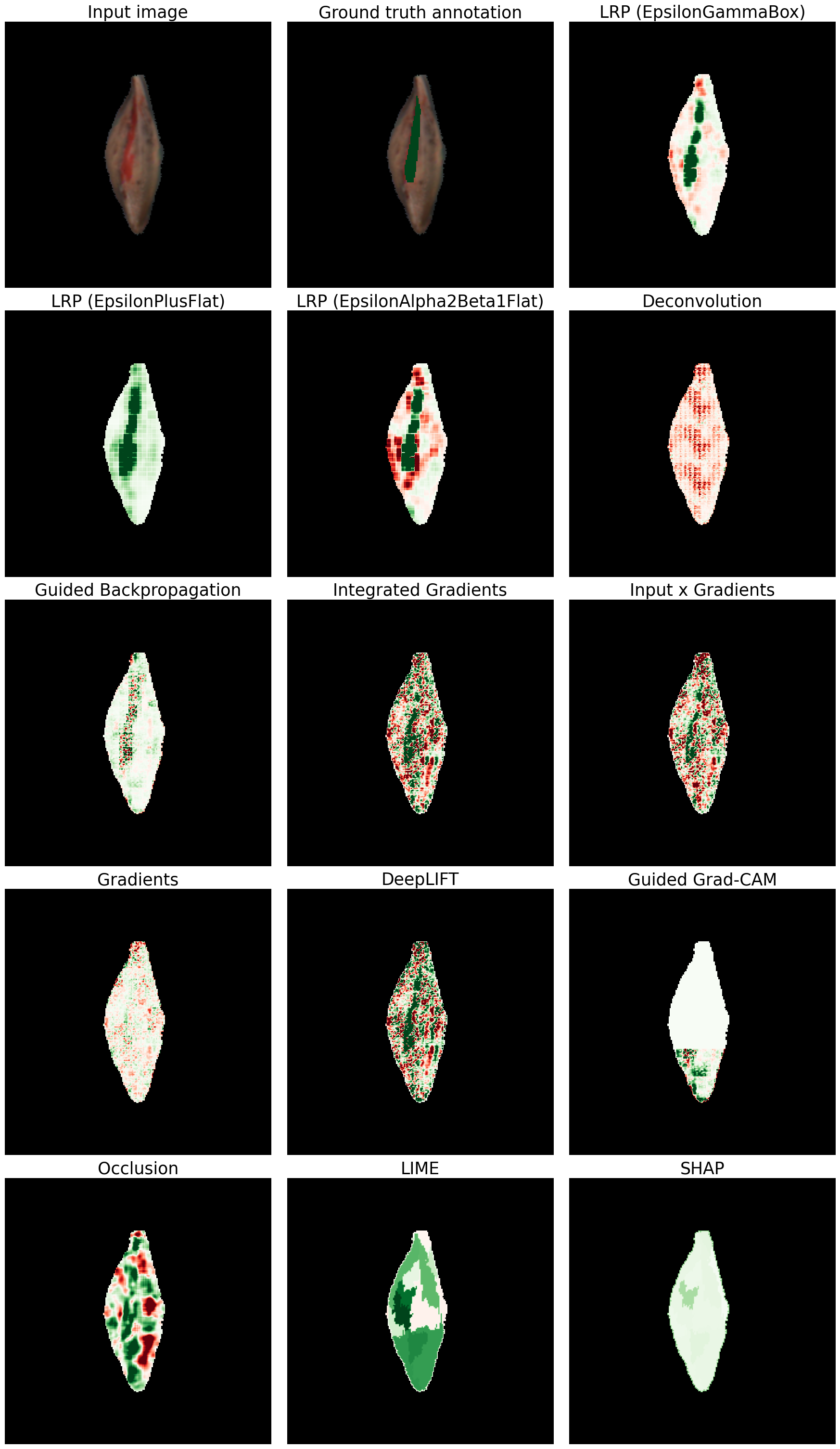}
    \caption{}
    \end{subfigure}
\end{figure}%
\begin{figure}[ht]\ContinuedFloat
\centering
    \begin{subfigure}[t]{0.7\linewidth}
    \includegraphics[width=\linewidth]{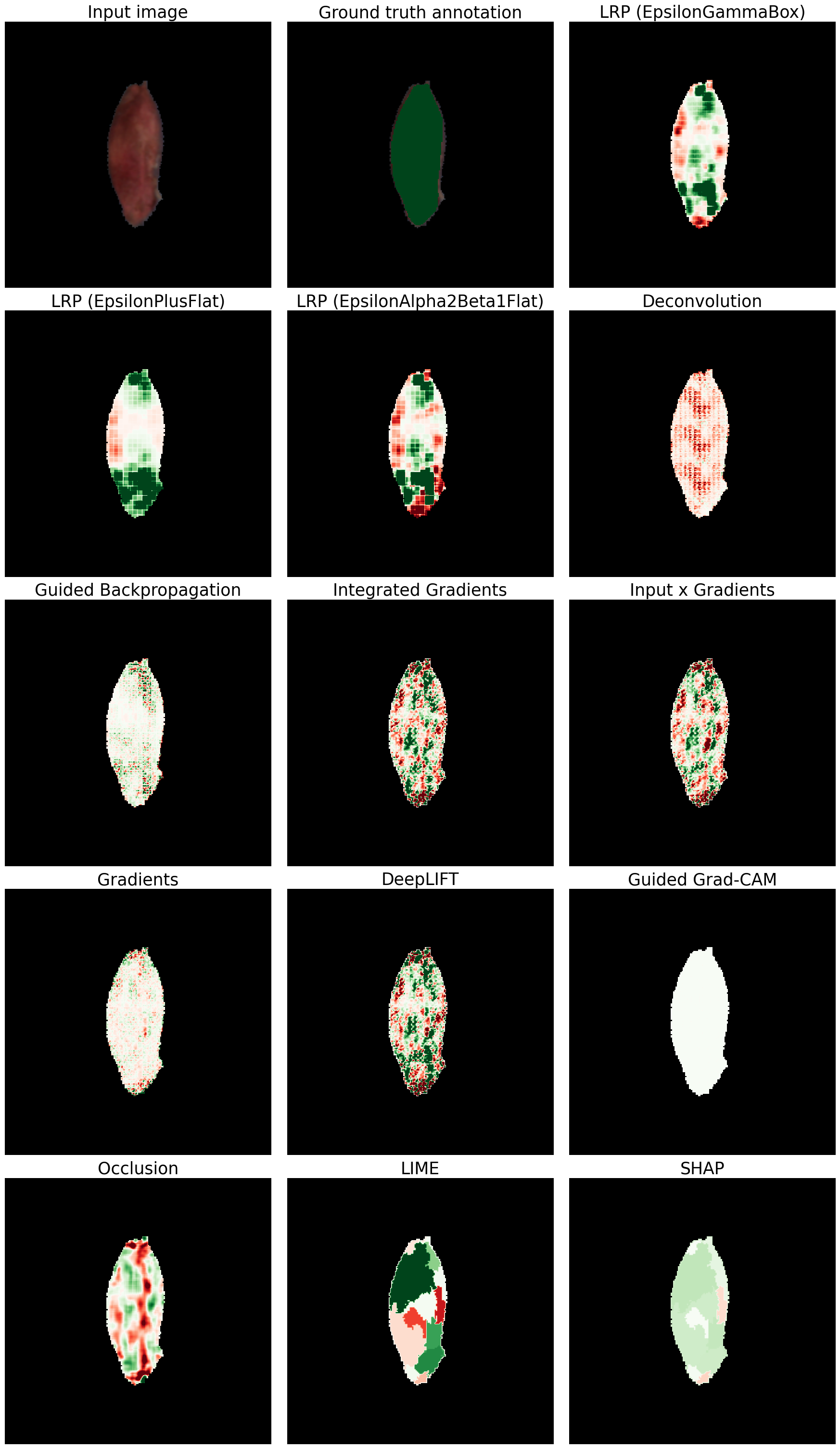}
    \caption{}
    \end{subfigure}
\end{figure}%
\begin{figure}[ht]\ContinuedFloat
\centering
    \begin{subfigure}[t]{0.7\linewidth}
    \includegraphics[width=\linewidth]{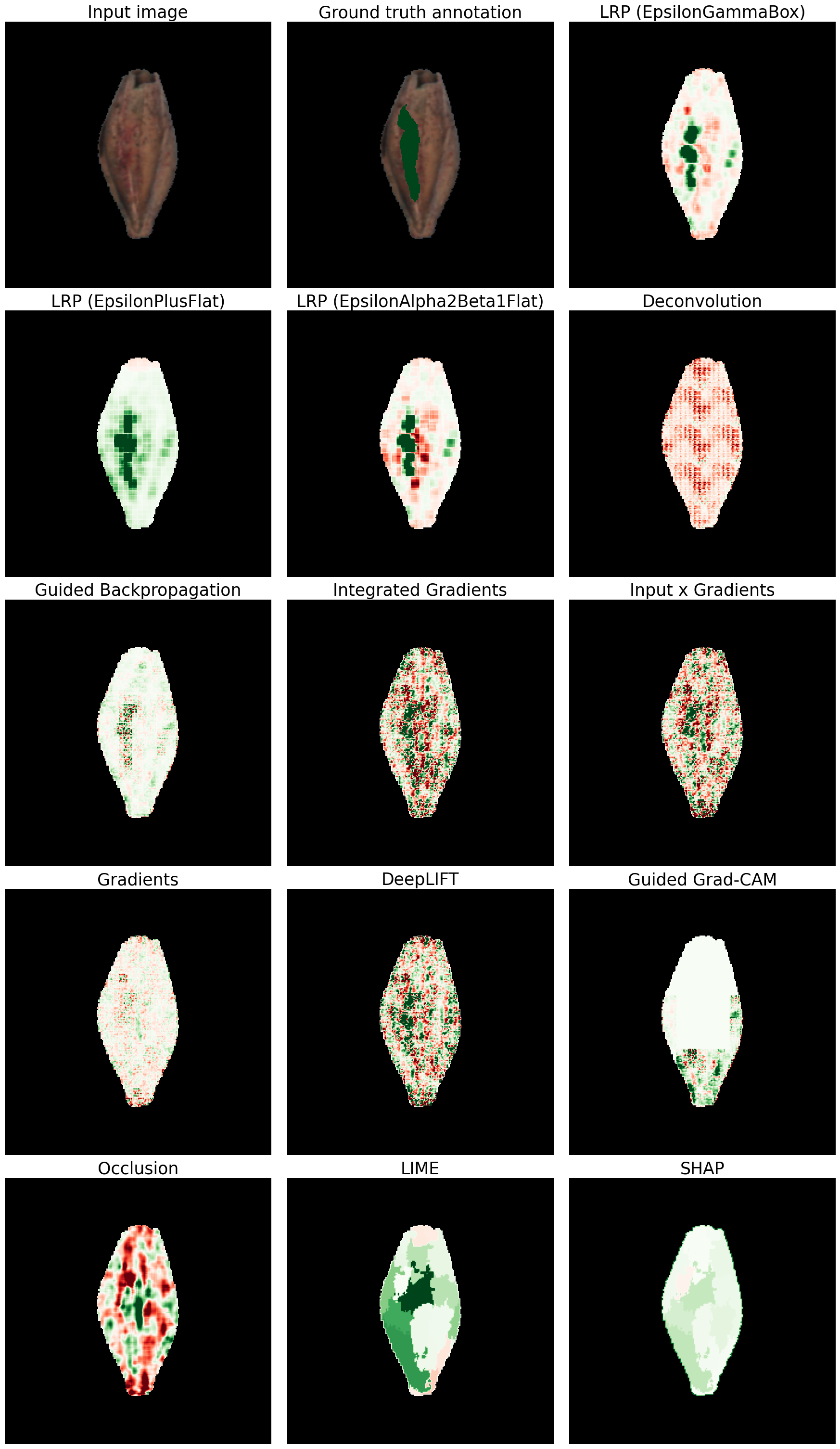}
    \caption{}
    \end{subfigure}
\end{figure}%
\begin{figure}[ht]\ContinuedFloat
\centering
    \begin{subfigure}[t]{0.7\linewidth}
    \includegraphics[width=\linewidth]{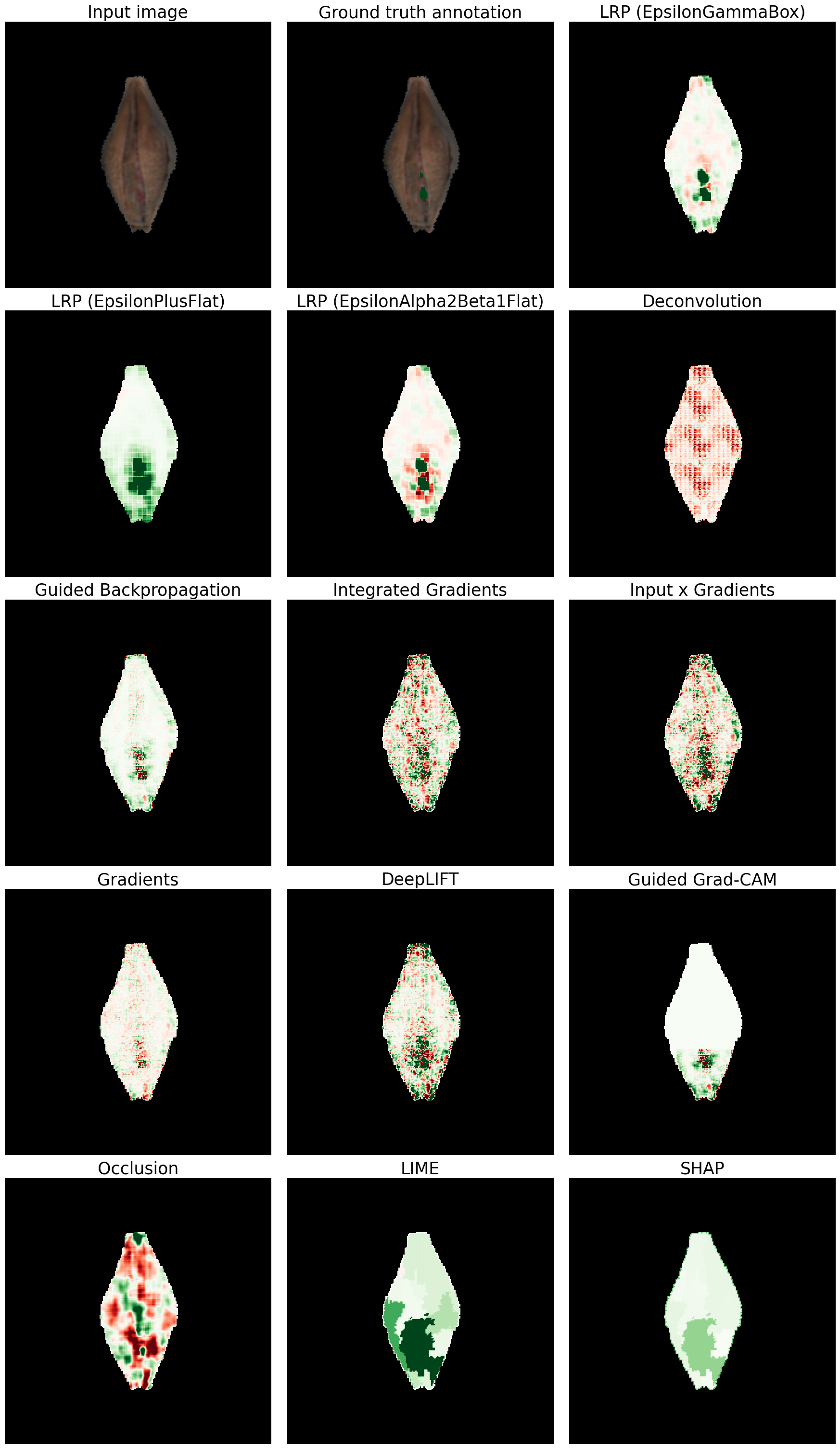}
    \caption{}
    \end{subfigure}
\end{figure}%
\begin{figure}[ht]\ContinuedFloat
\centering
    \begin{subfigure}[t]{0.7\linewidth}
    \includegraphics[width=\linewidth]{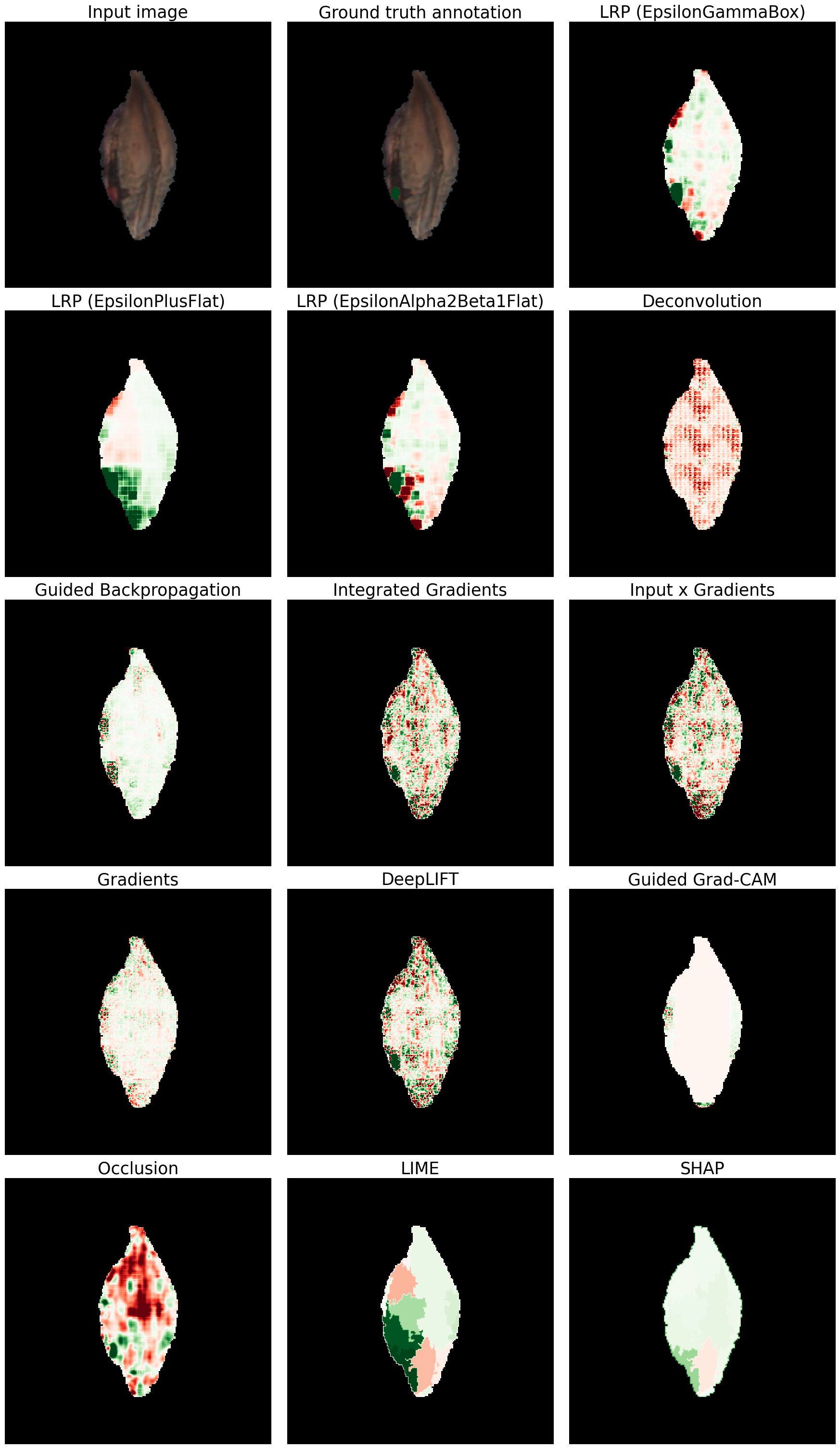}
    \caption{}
    \end{subfigure}
    \caption{All explanations -- pink fusarium}
    \label{fig:PF_examples}
\end{figure}

\subsection{Skinned}
\autoref{fig:Skinned_examples} shows examples of grains with a damage "skinned", ground truth annotations and explanations from all methods used in this paper.

\begin{figure}[htbp]
\centering
    \begin{subfigure}[t]{0.7\linewidth}
    \includegraphics[width=\linewidth]{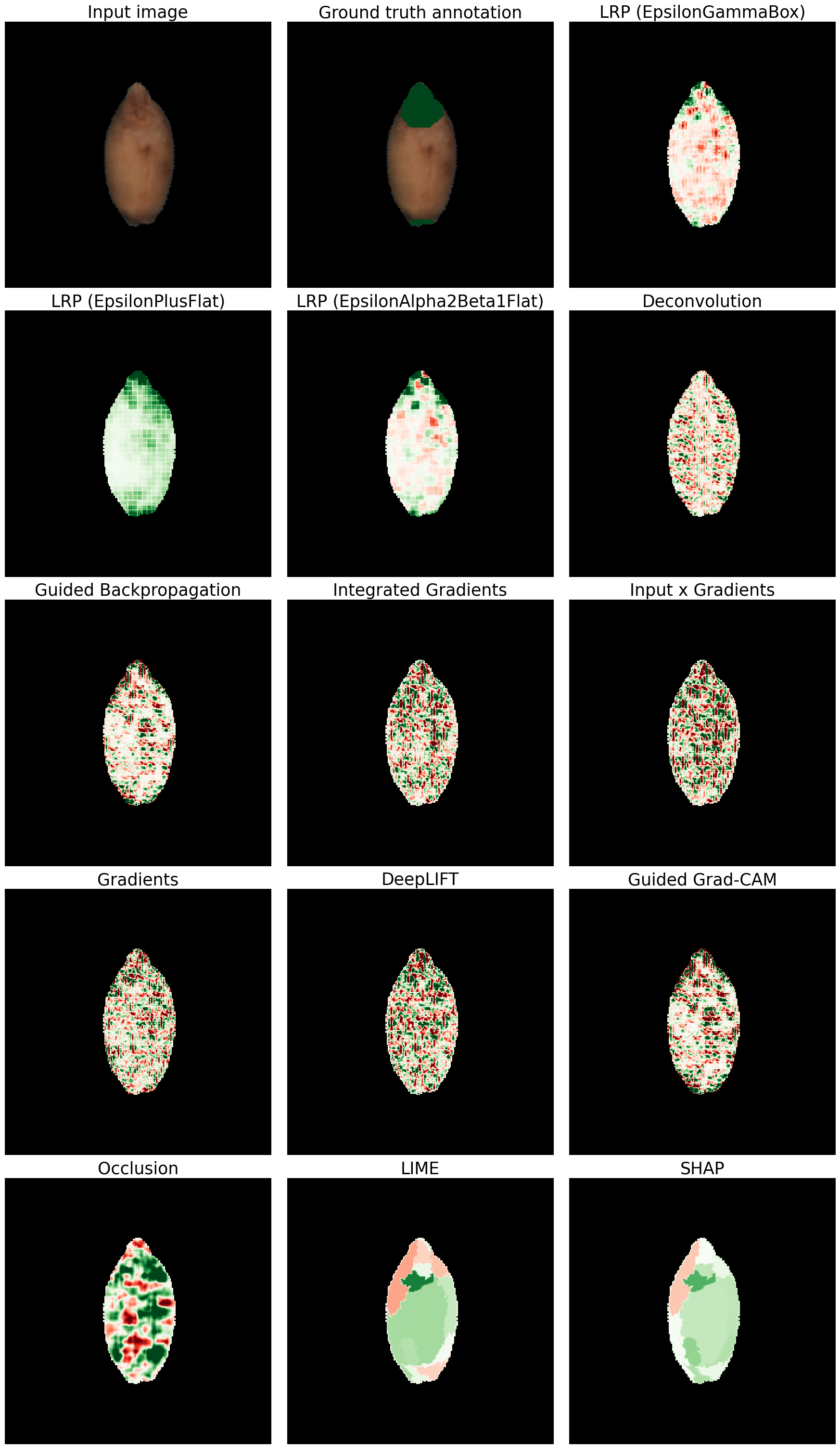}
    \caption{}
    \end{subfigure}
\end{figure}%
\begin{figure}[ht]\ContinuedFloat
\centering
    \begin{subfigure}[t]{0.7\linewidth}
    \includegraphics[width=\linewidth]{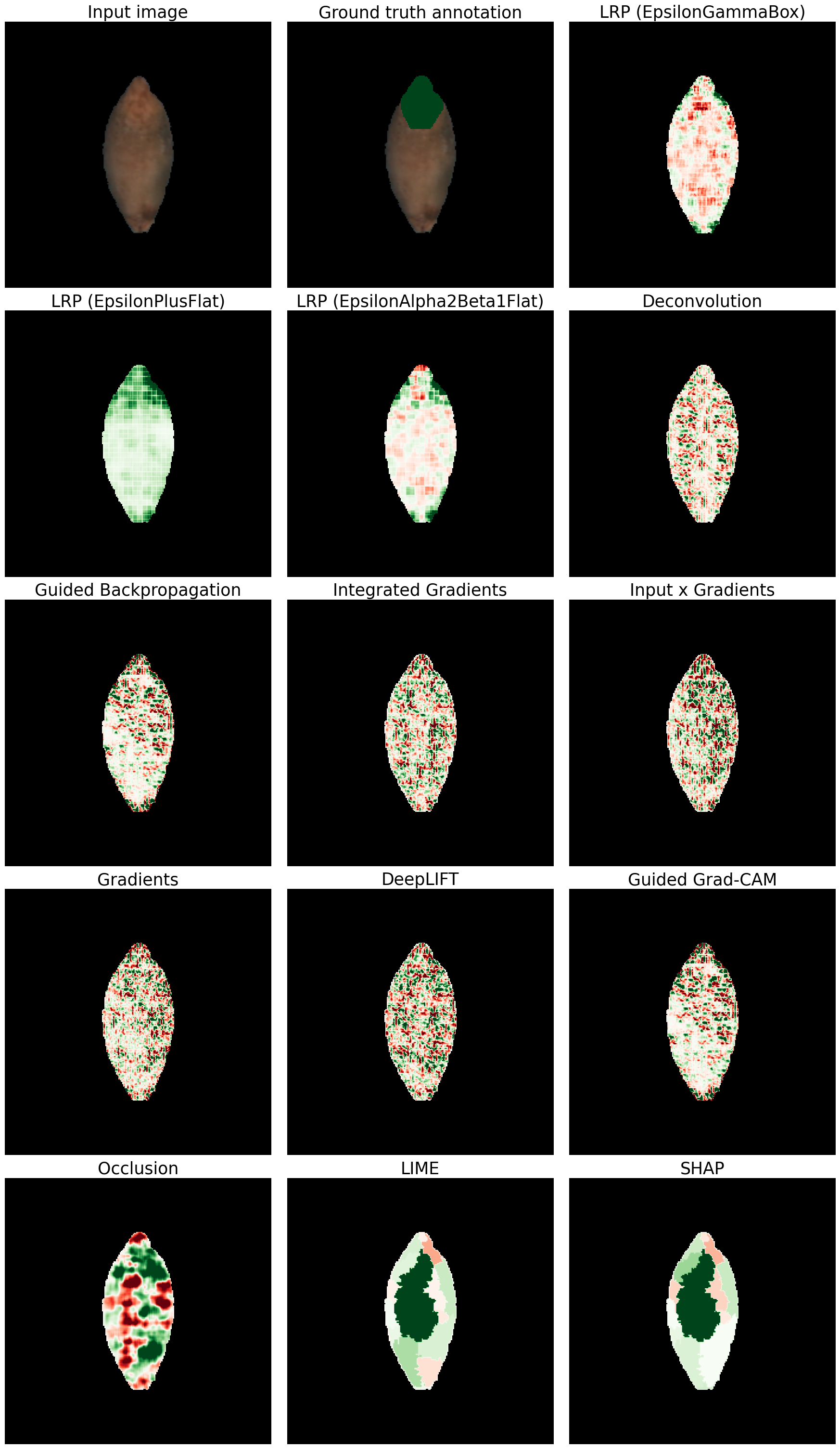}
    \caption{}
    \end{subfigure}
\end{figure}%
\begin{figure}[ht]\ContinuedFloat
\centering
    \begin{subfigure}[t]{0.7\linewidth}
    \includegraphics[width=\linewidth]{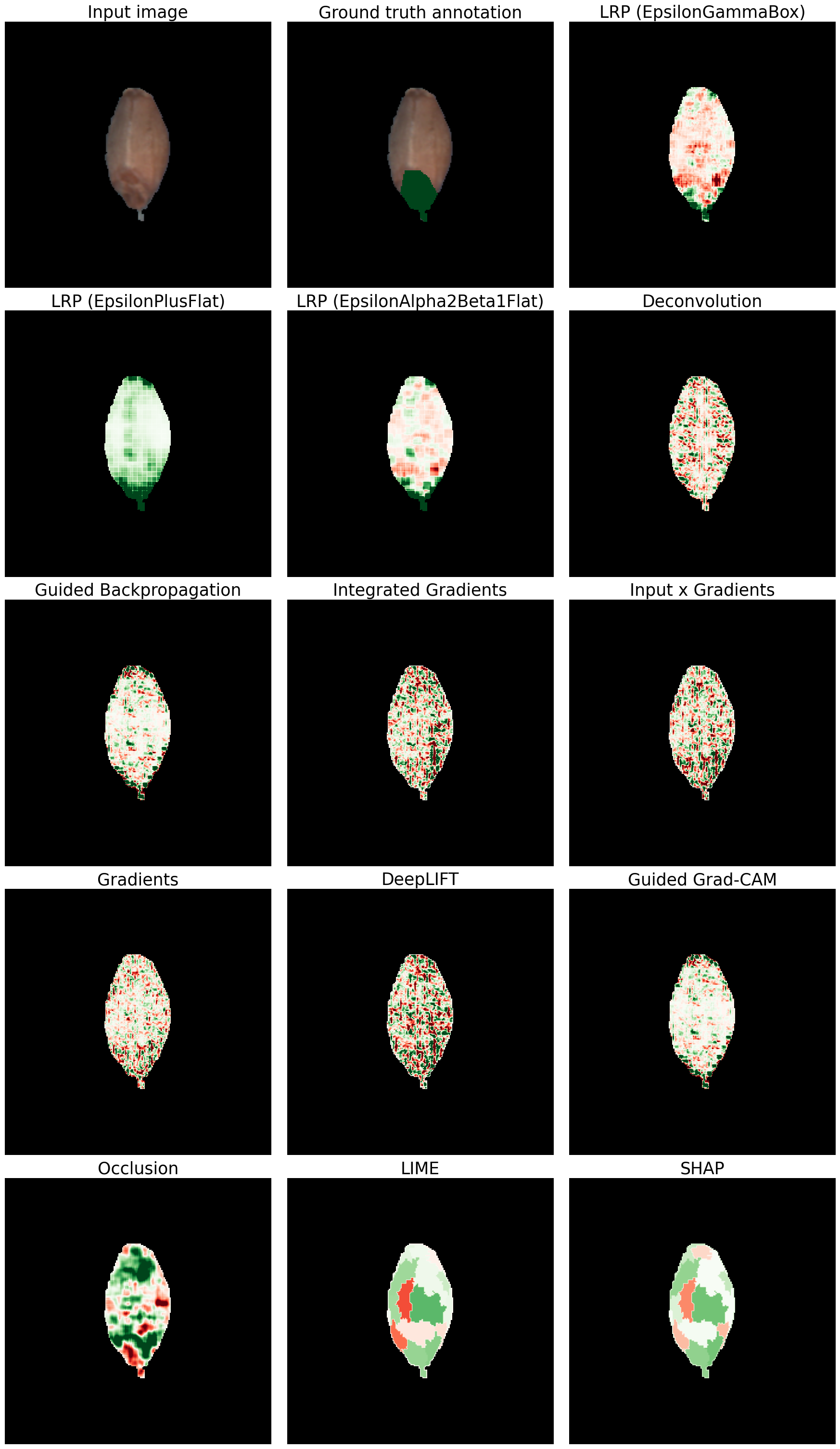}
    \caption{}
    \end{subfigure}
\end{figure}%
\begin{figure}[ht]\ContinuedFloat
 \centering
   \begin{subfigure}[t]{0.7\linewidth}
    \includegraphics[width=\linewidth]{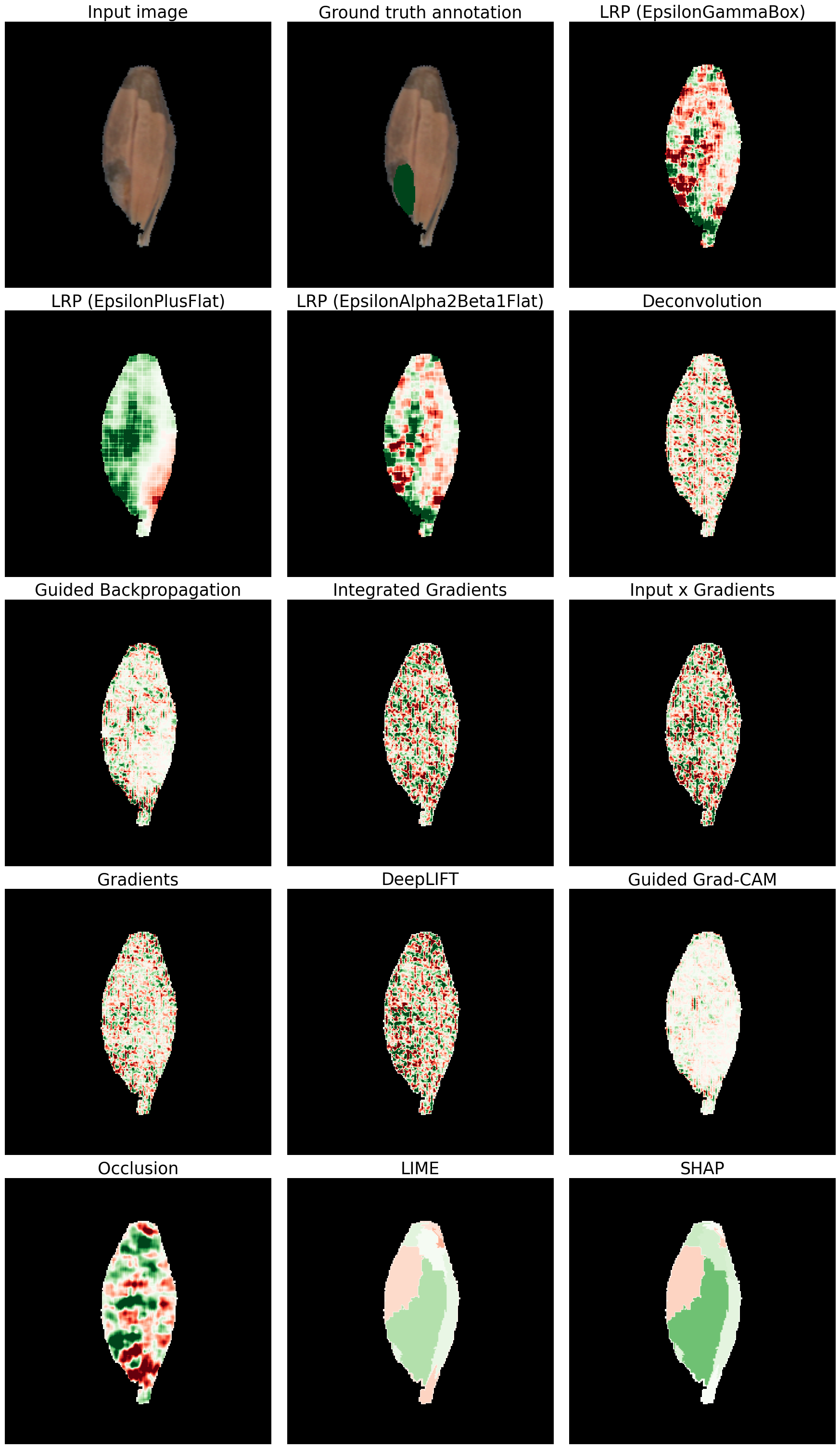}
    \caption{}
    \end{subfigure}
\end{figure}%
\begin{figure}[ht]\ContinuedFloat
\centering
    \begin{subfigure}[t]{0.7\linewidth}
    \includegraphics[width=\linewidth]{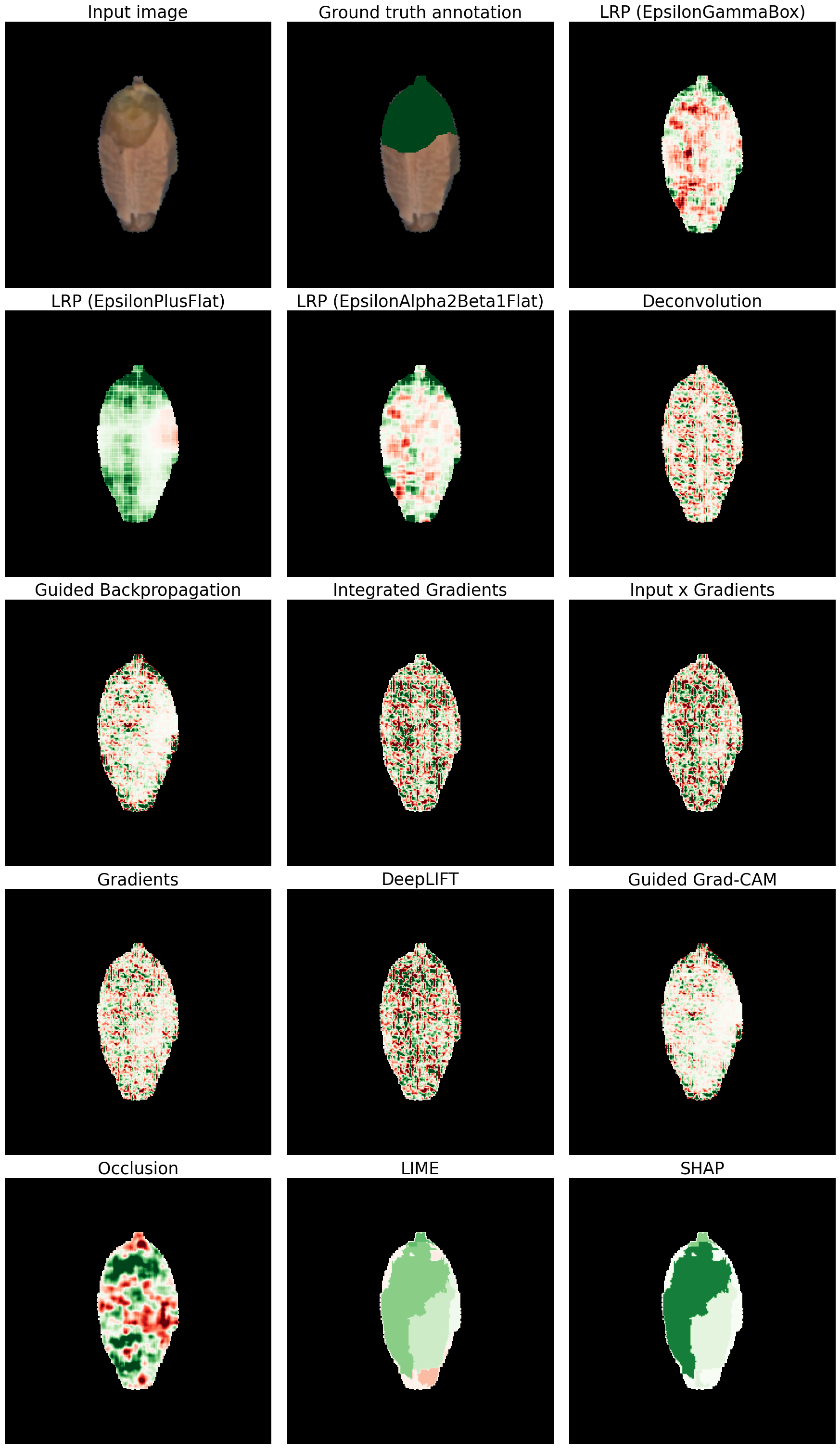}
    \caption{}
    \end{subfigure}
    \caption{All explanations -- skinned}
    \label{fig:Skinned_examples}
\end{figure}

\section{Detailed results}\label{sec:tables}

\subsection{Pink Fusarium}
The detailed results can be found in the following tables: robustness to data augmentation in \autoref{tab:PinkFusarium_robustness}, pixel flipping in \autoref{tab:PixelFlipping_PinkFusarium}, IROF in \autoref{tab:IROF_PinkFusarium}, sensitivity and complexity in \autoref{tab:quality_PinkFusarium}, ROC-AUC in \autoref{tab:ROC_PinkFusarium}, and Relevance Mass Accuracy in \autoref{tab:PinkFusarium_RelevanceMassAccuracy}.

\begin{table}[htp]
  \centering
  \begin{tabular}{|l||l|l|l||l|l|l|} \hline
     & Brightness & Hue & Saturation & Rotate & Scale & Translate \\ 
     & [-10, 10] & [-5, 5] & [-55, 55] & [-40, 40] & [0.8, 1.2] & [-0.08, 0.08] \\
     \hline \hline
    Gradients 	 & 0.863$\pm$0.007 	 & 1.002$\pm$0.007 	 & 0.894$\pm$0.006 	 & 0.364$\pm$0.007 	 & 0.537$\pm$0.008 	 & 0.415$\pm$0.009 	 \\ \hline
Input x Gradients 	 & 0.909$\pm$0.011 	 & 0.907$\pm$0.010 	 & 0.783$\pm$0.010 	 & 0.408$\pm$0.013 	 & 0.623$\pm$0.013 	 & 0.398$\pm$0.012 	 \\ \hline
Integrated Gradients 	 & 0.973$\pm$0.011 	 & 0.978$\pm$0.010 	 & 0.903$\pm$0.010 	 & 0.444$\pm$0.016 	 & 0.692$\pm$0.013 	 & 0.418$\pm$0.015 	 \\ \hline
Deconvolution 	 & \textbf{1.059}$\pm$0.007 	 & \textbf{1.128}$\pm$0.008 	 & \textbf{1.045}$\pm$0.007 	 & 0.420$\pm$0.006 	 & 0.593$\pm$0.007 	 & 0.465$\pm$0.008 	 \\ \hline
Guided Backprop 	 & 0.947$\pm$0.009 	 & 1.069$\pm$0.007 	 & 0.979$\pm$0.006 	 & 0.630$\pm$0.013 	 & 0.710$\pm$0.012 	 & \textbf{0.660}$\pm$0.012	 \\ \hline
LRP  	 & 0.914$\pm$0.006 	 & 0.912$\pm$0.007 	 & 0.891$\pm$0.005 	 & \textbf{0.704}$\pm$0.009 	 & \textbf{0.819}$\pm$0.006 	 & \textbf{0.653}$\pm$0.012  	 \\ 
EpsilonPlusFlat & & & & & & \\ \hline
LRP  	 & 0.872$\pm$0.006 	 & 0.893$\pm$0.006 	 & 0.870$\pm$0.005 	 & 0.550$\pm$0.011 	 & 0.708$\pm$0.008 	 & 0.536$\pm$0.010 	 \\ 
EpsilonGammaBox & & & & & & \\ \hline
LRP 	 & 0.877$\pm$0.006 	 & 0.896$\pm$0.006 	 & 0.862$\pm$0.005 	 & 0.583$\pm$0.009 	 & 0.743$\pm$0.007 	 & 0.552$\pm$0.011 	 \\ 
 Epsilon $\alpha 2 \beta 1 $ Flat & & & & & & \\ \hline
DeepLIFT 	 & 0.961$\pm$0.011 	 & 1.000$\pm$0.010 	 & 0.902$\pm$0.009 	 & 0.515$\pm$0.013 	 & 0.710$\pm$0.014 	 & 0.499$\pm$0.012 	 \\ \hline
SHAP 	 & 0.863$\pm$0.007 	 & 1.002$\pm$0.007 	 & 0.894$\pm$0.006 	 & 0.364$\pm$0.007 	 & 0.537$\pm$0.008 	 & 0.415$\pm$0.009 	 \\ \hline
LIME 	 & 0.863$\pm$0.007 	 & 1.002$\pm$0.007 	 & 0.894$\pm$0.006 	 & 0.364$\pm$0.007 	 & 0.537$\pm$0.008 	 & 0.415$\pm$0.009 	 \\ \hline
Occlusion 	 & 0.864$\pm$0.014 	 & 0.970 $\pm$ 0.007 	 & - 	 & - 	 & - 	 & - 	 \\ \hline
  \end{tabular}
  \caption{Pink fusarium: Robustness $\uparrow$ to data augmentation as described in \autoref{sec:robustness}. Uncertainties are the standard errors of the mean (389 correctly classified images). The intervals for each augmentation method were chosen, so that the probability of the target class drops by at least $10\%$ at one of the endpoints. The first three methods are considered invariant and the latter ones are  equivariant. Values below $1$ mean that the explanations are less robust than the probabilities of the classifier, higher values signalize higher robustness of explanations when compared to the robustness of the model itself.}
  \label{tab:PinkFusarium_robustness}
\end{table}

\begin{table}[htp]
  \centering
  \begin{tabular}{|l||l|l|l|l|} \hline
                                & mean pooling      & max pooling       & max abs pooling   & $\ell_2$-norm pooling \\ \hline \hline
    Gradients                   & 0.697 $\pm$ 0.020 & 0.390 $\pm$ 0.017 & 0.304 $\pm$ 0.016 & 0.307 $\pm$ 0.016     \\ \hline 
    Input x Gradients           & \textbf{0.738} $\pm$ 0.020 & 0.393 $\pm$ 0.016 & 0.349 $\pm$ 0.017 & 0.349 $\pm$ 0.017     \\ \hline 
    Integrated Gradients        & \textbf{0.741} $\pm$ 0.019 & 0.412 $\pm$ 0.017 & 0.357 $\pm$ 0.017 & 0.357 $\pm$ 0.017     \\ \hline
    Guided Backprop             & 0.100 $\pm$ 0.010 & 0.388 $\pm$ 0.017 & \textbf{0.431} $\pm$ 0.019 & \textbf{0.432} $\pm$ 0.019     \\ \hline 
    Deconvolution               & 0.028 $\pm$ 0.006 & 0.063 $\pm$ 0.009 & 0.256 $\pm$ 0.016 & 0.260 $\pm$ 0.016     \\ \hline
    LRP: EpsilonPlusFlat        & 0.446 $\pm$ 0.019 & 0.446 $\pm$ 0.019 & 0.407 $\pm$ 0.018 & 0.409 $\pm$ 0.018     \\ \hline
    LRP: EpsilonGammaBox        & 0.492 $\pm$ 0.021 & \textbf{0.506} $\pm$ 0.021 & 0.385 $\pm$ 0.017 & 0.387 $\pm$ 0.017     \\ \hline
    LRP: EpsilonAlpha2Beta1Flat & 0.391 $\pm$ 0.021 & 0.392 $\pm$ 0.021 & 0.391 $\pm$ 0.017 & 0.391 $\pm$ 0.017     \\ \hline
    DeepLIFT                    & 0.697 $\pm$ 0.020 & 0.425 $\pm$ 0.017 & 0.356 $\pm$ 0.017 & 0.356 $\pm$ 0.017     \\ \hline 
    Guided Grad-CAM  & 0.079 $\pm$ 0.010     & 0.148 $\pm$ 0.013     & 0.150 $\pm$ 0.014     & 0.151 $\pm$ 0.014     \\ \hline 
    LIME & \multicolumn{2}{|c|}{0.406 $\pm$ 0.018} & \multicolumn{2}{|c|}{0.336 $\pm$ 0.017} \\ \hline 
    SHAP & \multicolumn{2}{|c|}{0.091 $\pm$ 0.010} & \multicolumn{2}{|c|}{0.036 $\pm$ 0.008}    \\ \hline 
    Occlusion & \multicolumn{2}{|c|}{0.490 $\pm$ 0.021}  & \multicolumn{2}{|c|}{0.365 $\pm$ 0.016}  \\ \hline \hline
    mean                        & \textbf{0.727} $\pm$ 0.019 & \textbf{0.506} $\pm$ 0.017 & 0.393 $\pm$ 0.018 & 0.393 $\pm$ 0.018     \\ \hline 
    ground-truth & \multicolumn{4}{|c|}{0.180 $\pm$ 0.016} \\ \hline
  \end{tabular}
  \caption{Pink fusarium: pixel-flipping $\uparrow$ for each explainability method. Uncertainties are the standard error of the mean (389 images).}
  \label{tab:PixelFlipping_PinkFusarium}
\end{table}

\begin{table}[htp]
  \centering
  \begin{tabular}{|l||l|l|l|l|} \hline
                                & Mean              & Max               & Max abs           & $\ell_2$-norm     \\ \hline \hline
    Gradients                   & 0.214 $\pm$ 0.012 & 0.224 $\pm$ 0.013 & 0.218 $\pm$ 0.013 & 0.219 $\pm$ 0.013 \\ \hline 
    Input x Gradients           & 0.258 $\pm$ 0.014 & 0.216 $\pm$ 0.012 & 0.231 $\pm$ 0.013 & 0.233 $\pm$ 0.013 \\ \hline 
    Integrated Gradients        & 0.248 $\pm$ 0.013 & 0.225 $\pm$ 0.012 & 0.239 $\pm$ 0.013 & 0.240 $\pm$ 0.013 \\ \hline
    Guided Backprop             & 0.135 $\pm$ 0.007 & 0.257 $\pm$ 0.013 & 0.274 $\pm$ 0.014 & 0.274 $\pm$ 0.014 \\ \hline 
    Deconvolution               & 0.197 $\pm$ 0.014 & 0.163 $\pm$ 0.009 & 0.225 $\pm$ 0.012 & 0.223 $\pm$ 0.012 \\ \hline
    LRP: EpsilonPlusFlat        & 0.282 $\pm$ 0.015 & 0.282 $\pm$ 0.015 & 0.282 $\pm$ 0.015 & 0.280 $\pm$ 0.014 \\ \hline
    LRP: EpsilonGammaBox        & 0.273 $\pm$ 0.014 & 0.276 $\pm$ 0.013 & 0.262 $\pm$ 0.014 & 0.260 $\pm$ 0.014 \\ \hline
    LRP: EpsilonAlpha2Beta1Flat & 0.253 $\pm$ 0.013 & 0.254 $\pm$ 0.013 & 0.268 $\pm$ 0.014 & 0.268 $\pm$ 0.014 \\ \hline
    DeepLIFT                    & 0.252 $\pm$ 0.013 & 0.223 $\pm$ 0.012 & 0.226 $\pm$ 0.012 & 0.227 $\pm$ 0.013 \\ \hline 
    Guided Grad-CAM    & 0.143 $\pm$ 0.009     & 0.156 $\pm$ 0.010     & 0.152 $\pm$ 0.011     & 0.152 $\pm$ 0.011     \\ \hline
    LIME & \multicolumn{2}{|c|}{0.376 $\pm$ 0.017} & \multicolumn{2}{|c|}{0.304 $\pm$ 0.016} \\ \hline 
    SHAP & \multicolumn{2}{|c|}{\textbf{0.475} $\pm$ 0.019} & \multicolumn{2}{|c|}{\textbf{0.369} $\pm$ 0.021} \\ \hline 
    Occlusion  & \multicolumn{2}{|c|}{0.232 $\pm$ 0.013}     & \multicolumn{2}{|c|}{0.240 $\pm$ 0.014 }     \\ \hline   \hline
    mean                        & 0.361 $\pm$ 0.015 & 0.367 $\pm$ 0.016 & 0.296 $\pm$ 0.016 & 0.297 $\pm$ 0.016 \\ \hline
    %ground-truth & \multicolumn{4}{|c|}{} \\ \hline
  \end{tabular}
   \caption{PinkFusarium: IROF $\uparrow$ - different channel poolings. Uncertainties are the standard error of the mean (389 images).}
  \label{tab:IROF_PinkFusarium}
\end{table}

\begin{table}[htp]
  \centering
  \begin{tabular}{|l||l|l|l|l|} \hline
                                & Average Sensitivity  $\downarrow$     & Complexity $\downarrow$ \\ \hline \hline
    Gradients        & 4.174 $\pm$ 0.118     & 8.822 $\pm$ 0.022     \\ \hline      
    Input x Gradients   & 2.373 $\pm$ 0.068     & 8.835 $\pm$ 0.022     \\ \hline 
    Integrated Gradients      & 2.730 $\pm$ 0.090     & 8.868 $\pm$ 0.022     \\ \hline 
    Deconvolution    & 2.237 $\pm$ 0.034     & 8.796 $\pm$ 0.023     \\ \hline              
    Guided Backpropagation   & 12.194 $\pm$ 0.691    & 8.337 $\pm$ 0.023     \\ \hline                             
    LRP: EpsilonPlusFlat   & 1.443 $\pm$ 0.026     & 8.766 $\pm$ 0.027     \\ \hline     LRP: EpsilonGammaBox   & \textbf{1.347} $\pm$ 0.020     & 8.554 $\pm$ 0.027     \\ \hline 
    LRP: EpsilonAlpha2Beta1Flat    & \textbf{1.330} $\pm$ 0.020     & 8.498 $\pm$ 0.029     \\ \hline
    DeepLIFT         & 4.312 $\pm$ 0.134     & 8.747 $\pm$ 0.022     \\ \hline
    Guided Grad-CAM    & 16.595 $\pm$ 14.708   & \textbf{7.006} $\pm$ 0.434     \\ \hline 
    Occlusion        & 2.611 $\pm$ 0.072     & 9.126 $\pm$ 0.023     \\ \hline      
  \end{tabular}
  \caption{PinkFusarium: Evaluation of quality of the explanations. Uncertainties are the standard errors of the mean (389 images).}
  \label{tab:quality_PinkFusarium}
\end{table}

\begin{table}[htp]
  \centering
  \begin{tabular}{|l||l|l|l|l|} \hline
                                & mean pooling      & max pooling       & max abs pooling   & $\ell_2$-norm pooling \\ \hline \hline
    Gradients                   & 0.731 $\pm$ 0.001 & 0.958 $\pm$ 0.001 & 0.955 $\pm$ 0.002 & 0.956 $\pm$ 0.002     \\ \hline 
    Input x Gradients           & 0.756 $\pm$ 0.002 & 0.963 $\pm$ 0.001 & 0.973 $\pm$ 0.001 & 0.973 $\pm$ 0.001     \\ \hline 
    Integrated Gradients        & 0.765 $\pm$ 0.003 & 0.963 $\pm$ 0.001 & 0.975 $\pm$ 0.001 & 0.975 $\pm$ 0.001     \\ \hline
    Guided Backpropagation             & 0.787 $\pm$ 0.001 & 0.973 $\pm$ 0.001 & 0.979 $\pm$ 0.001 & 0.979 $\pm$ 0.001     \\ \hline 
    Deconvolution               & 0.637 $\pm$ 0.001 & 0.816 $\pm$ 0.002 & 0.858 $\pm$ 0.002 & 0.858 $\pm$ 0.002     \\ \hline
    LRP: EpsilonPlusFlat        & \textbf{0.965}. $\pm$ 0.005 & \textbf{0.965} $\pm$ 0.005 & \textbf{0.982} $\pm$ 0.001 & \textbf{0.982} $\pm$ 0.001     \\ \hline
    LRP: EpsilonGammaBox        & 0.848 $\pm$ 0.006 & 0.877 $\pm$ 0.005 & \textbf{0.983} $\pm$ 0.001 & \textbf{0.983} $\pm$ 0.001     \\ \hline
    LRP: EpsilonAlpha2Beta1Flat & 0.825 $\pm$ 0.006 & 0.825 $\pm$ 0.006 & 0.980 $\pm$ 0.001 & 0.980 $\pm$ 0.001     \\ \hline
    DeepLIFT                    & 0.781 $\pm$ 0.002 & 0.966 $\pm$ 0.001 & 0.976 $\pm$ 0.001 & 0.976 $\pm$ 0.001     \\ \hline 
    GuidedGradCam    & 0.715 $\pm$ 0.011     & 0.751 $\pm$ 0.013     & 0.731 $\pm$ 0.014     & 0.731 $\pm$ 0.014     \\ \hline                                                
    LIME  & \multicolumn{2}{|c|}{0.901 $\pm$ 0.007} & \multicolumn{2}{|c|}{0.963 $\pm$ 0.003} \\ \hline 
    SHAP  & \multicolumn{2}{|c|}{0.592 $\pm$ 0.012} & \multicolumn{2}{|c|}{0.592 $\pm$ 0.012}   \\ \hline
    Occlusion        & \multicolumn{2}{|c|}{0.726 $\pm$ 0.006} & \multicolumn{2}{|c|}{0.966 $\pm$ 0.001}     \\ \hline  \hline
    mean                        & 0.820 $\pm$ 0.004 & \textbf{0.964} $\pm$ 0.002 & \textbf{0.983} $\pm$ 0.001 & \textbf{0.983} $\pm$ 0.001     \\ \hline 
  \end{tabular}
  \caption{Pink fusarium: ROC-AUC $\uparrow$ for each explainability method and channel pooling type. Uncertainties are the standard errors of the mean (173 images).}
  \label{tab:ROC_PinkFusarium}
\end{table}

\begin{table}
  \centering
  \begin{tabular}{|l||l|l|} \hline
                                & max abs pooling   & $\ell_2$-norm pooling \\ \hline \hline
    Gradients        & 0.125 $\pm$ 0.005     & 0.128 $\pm$ 0.005     \\ \hline
    Input x Gradients  & 0.373 $\pm$ 0.015     & 0.380 $\pm$ 0.015     \\ \hline
    Integrated Gradients  & 0.402 $\pm$ 0.016     & 0.409 $\pm$ 0.016    \\ \hline         
    Deconvolution    & 0.050 $\pm$ 0.003     & 0.050 $\pm$ 0.003     \\ \hline  
    Guided Backpropagation  & 0.405 $\pm$ 0.010     & 0.412 $\pm$ 0.011     \\ \hline      
    LRP: EpsilonPlusFlat   & \textbf{0.485} $\pm$ 0.014     & \textbf{0.485} $\pm$ 0.014    \\ \hline        
    LRP: EpsilonGammaBox   & 0.469 $\pm$ 0.012     & 0.458 $\pm$ 0.012    \\ \hline        
    LRP: EpsilonAlpha2Beta1Flat  & 0.551 $\pm$ 0.013     & 0.551 $\pm$ 0.013     \\ \hline 
    DeepLIFT   & 0.428 $\pm$ 0.015     & 0.437 $\pm$ 0.015     \\ \hline                   
    Guided Grad-CAM    & 0.143 $\pm$ 0.143     & 0.146 $\pm$ 0.146     \\ \hline \hline 
    mean     & 0.350 $\pm$ 0.017     & 0.350 $\pm$ 0.017     \\ \hline 
     
  \end{tabular}
  \caption{Pink fusarium: Relevance Mass Accuracy for each explainability method. Uncertainties are the standard error of the mean (173 images). Only max abs and $\ell_2$-norm pooling are included since this method expects non-negative explanations.}
  \label{tab:PinkFusarium_RelevanceMassAccuracy}
\end{table}

\subsection{Skinned}
The detailed results can be found in the following tables: robustness to data augmentation in \autoref{tab:Skinned_robustness}, pixel flipping in \autoref{tab:PixelFlipping_Skinned}, IROF in \autoref{tab:IROF_Skinned}, sensitivity and complexity in \autoref{tab:quality_skinned}, ROC-AUC in \autoref{tab:ROC_Skinned}, and Relevance Mass Accuracy in \autoref{tab:RelevanceMassAccuracy_Skinned}.

\begin{table}[htp]
  \centering
  \begin{tabular}{|l||l|l|l||l|l|l|} \hline
     & Brightness & Hue & Saturation  & Rotate & Scale & Translate \\
     & [-20, 20] & [-15, 15] & [-50, 50] & [-8, 8] & [0.8, 1.2] & [-0.04, 0.04] \\\hline \hline
     Gradients 	 & 0.838$\pm$0.007 	 & 0.883$\pm$0.006 	 & 0.871$\pm$0.006 	 & 0.317$\pm$0.003 	 & 0.249$\pm$0.004 	 & 0.378$\pm$0.005 	 \\ \hline
Input x Gradients 	 & 0.863$\pm$0.011 	 & 0.807$\pm$0.009 	 & 0.805$\pm$0.008 	 & 0.405$\pm$0.007 	 & 0.345$\pm$0.006 	 & 0.422$\pm$0.008 	 \\ \hline
Integrated Gradients 	 & 1.004$\pm$0.011 	 & 0.931$\pm$0.009 	 & 0.950$\pm$0.009 	 & 0.448$\pm$0.007 	 & 0.405$\pm$0.007 	 & 0.429$\pm$0.007 	 \\ \hline
Deconvolution 	 & \textbf{1.050}$\pm$0.008 	 & \textbf{1.037}$\pm$0.007 	 & \textbf{1.034}$\pm$0.007 	 & 0.333$\pm$0.003 	 & 0.277$\pm$0.003 	 & 0.481$\pm$0.006 	 \\ \hline
Guided Backprop 	 & 0.899$\pm$0.007 	 & 0.935$\pm$0.006 	 & 0.923$\pm$0.006 	 & 0.485$\pm$0.005 	 & 0.445$\pm$0.006 	 & 0.557$\pm$0.007 	 \\ \hline
LRP  	 & 0.927$\pm$0.006 	 & 0.953$\pm$0.005 	 & 0.953$\pm$0.005 	 & \textbf{0.718}$\pm$0.010 	 & \textbf{0.701}$\pm$0.010 	 & \textbf{0.632}$\pm$0.012 	 \\ 
EpsilonPlusFlat & & & & & & \\\hline
LRP  	 & 0.889$\pm$0.005 	 & 0.927$\pm$0.004 	 & 0.923$\pm$0.004 	 & 0.471$\pm$0.005 	 & 0.408$\pm$0.005 	 & 0.419$\pm$0.006 	 \\ 
EpsilonGammaBox & & & & & & \\ \hline
LRP 	 & 0.910$\pm$0.006 	 & 0.938$\pm$0.005 	 & 0.934$\pm$0.005 	 & 0.587$\pm$0.008 	 & 0.552$\pm$0.009 	 & 0.498$\pm$0.009 	 \\ 
EpsilonAlpha2Beta1Flat & & & & & & \\\hline
DeepLIFT 	 & 0.945$\pm$0.011 	 & 0.907$\pm$0.009 	 & 0.915$\pm$0.008 	 & 0.415$\pm$0.006 	 & 0.363$\pm$0.005 	 & 0.373$\pm$0.006 	 \\ \hline
SHAP 	 & 0.838$\pm$0.007 	 & 0.883$\pm$0.006 	 & 0.871$\pm$0.006 	 & 0.317$\pm$0.003 	 & 0.249$\pm$0.004 	 & 0.378$\pm$0.005 	 \\ \hline
LIME 	 & 0.838$\pm$0.007 	 & 0.883$\pm$0.006 	 & 0.871$\pm$0.006 	 & 0.317$\pm$0.003 	 & 0.249$\pm$0.004 	 & 0.378$\pm$0.005 	 \\ \hline
Occlusion 	 & 0.930$\pm$0.011 	 & - 	 & - 	 & - 	 & - 	 & - 	 \\ \hline
    \end{tabular}
  \caption{Skinned: Robustness $\uparrow$ to data augmentationas described in \autoref{sec:robustness}. Uncertainties are the standard errors of the mean (383 correctly classified images). The intervals for each augmentation method were chosen, so that the probability of the target class drops by at least $10\%$ at one of the endpoints. The first three methods are considered invariant and the latter ones are equivariant. Values below $1$ mean that the explanations are less robust than the probabilities of the classifier, higher values signalize higher robustness of explanations when compared to the robustness of the model itself.}
  \label{tab:Skinned_robustness}
\end{table}

\begin{table}[htp]
  \centering
  \begin{tabular}{|l||l|l|l|l|} \hline
                                & mean pooling      & max pooling       & max abs pooling   & $\ell_2$-norm pooling \\ \hline \hline
    Gradients                   & 0.519 $\pm$ 0.025 & 0.504 $\pm$ 0.024 & 0.538 $\pm$ 0.017 & 0.548 $\pm$ 0.017 \\ \hline 
    Input x Gradients           & 0.503 $\pm$ 0.025 & 0.497 $\pm$ 0.025 & 0.503 $\pm$ 0.008 & 0.529 $\pm$ 0.008 \\ \hline 
    Integrated Gradients        & 0.495 $\pm$ 0.025 & 0.482 $\pm$ 0.025 & 0.528 $\pm$ 0.008 & 0.528 $\pm$ 0.008 \\ \hline
    Guided Backpropagation             & 0.507 $\pm$ 0.024 & 0.488 $\pm$ 0.023 & 0.557 $\pm$ 0.018 & 0.560 $\pm$ 0.018  \\ \hline 
    Deconvolution               & 0.449 $\pm$ 0.022 & 0.433 $\pm$ 0.021 & 0.449 $\pm$ 0.021 & 0.454 $\pm$ 0.021 \\ \hline
    LRP: EpsilonPlusFlat        & 0.658 $\pm$ 0.013 & 0.658 $\pm$ 0.013 & \textbf{0.605} $\pm$ 0.013 & \textbf{0.606} $\pm$ 0.013 \\ \hline
    LRP: EpsilonGammaBox        & 0.573 $\pm$ 0.022 & 0.568 $\pm$ 0.022 & \textbf{0.610} $\pm$ 0.015 & \textbf{0.607} $\pm$ 0.015 \\ \hline
    LRP: EpsilonAlpha2Beta1Flat & 0.571 $\pm$ 0.022 & 0.571 $\pm$ 0.022 & 0.568 $\pm$ 0.021 & 0.570 $\pm$ 0.020 \\ \hline
    DeepLIFT                    & 0.498 $\pm$ 0.025 & 0.491 $\pm$ 0.025 & 0.522 $\pm$ 0.008 & 0.520 $\pm$ 0.008 \\ \hline 
    Guided Grad-CAM  & 0.484 $\pm$ 0.023     & 0.465 $\pm$ 0.023     & 0.533 $\pm$ 0.018     & 0.535 $\pm$ 0.018     \\ \hline  
    LIME & \multicolumn{2}{|c|}{\textbf{0.835} $\pm$ 0.011} & \multicolumn{2}{|c|}{0.563 $\pm$ 0.015} \\ \hline 
    SHAP & \multicolumn{2}{|c|}{0.755 $\pm$ 0.015} & \multicolumn{2}{|c|}{0.545 $\pm$ 0.017} \\ \hline 
    Occlusion & \multicolumn{2}{|c|}{0.681 $\pm$ 0.019} & \multicolumn{2}{|c|}{0.522 $\pm$ 0.009}     \\ \hline   \hline
    mean                        & 0.497 $\pm$ 0.024 & 0.517 $\pm$ 0.023 & 0.555 $\pm$ 0.009 & 0.554 $\pm$ 0.009 \\ \hline 
    ground-truth & \multicolumn{4}{|c|}{0.499 $\pm$ 0.018} \\ \hline
  \end{tabular}
  \caption{Skinned: pixel-flipping $\uparrow$ for each explainability method and channel pooling type. Uncertainties are the standard error of the mean (383 images).}
  \label{tab:PixelFlipping_Skinned}
\end{table}

\begin{table}
  \centering
  \begin{tabular}{|l||l|l|l|l|} \hline
                                & Mean                          & Max               & Max abs           & $\ell_2$-norm     \\ \hline \hline
    Gradients                   & 0.365 $\pm$ 0.011             & 0.322 $\pm$ 0.010 & 0.314 $\pm$ 0.010 & 0.313 $\pm$ 0.010 \\ \hline 
    Input x Gradients           & 0.377 $\pm$ 0.012             & 0.333 $\pm$ 0.011 & 0.355 $\pm$ 0.011 & 0.352 $\pm$ 0.011 \\ \hline 
    Integrated Gradients        & 0.316 $\pm$ 0.011             & 0.328 $\pm$ 0.010 & 0.382 $\pm$ 0.011 & 0.375 $\pm$ 0.011 \\ \hline
    Guided Backpropagation             & 0.288 $\pm$ 0.010             & 0.318 $\pm$ 0.011 & 0.354 $\pm$ 0.011 & 0.351 $\pm$ 0.011 \\ \hline 
    Deconvolution               & 0.269 $\pm$ 0.010             & 0.374 $\pm$ 0.010 & 0.399 $\pm$ 0.011 & 0.394 $\pm$ 0.011 \\ \hline
    LRP: EpsilonPlusFlat        & 0.408 $\pm$ 0.011             & 0.408 $\pm$ 0.011 & 0.385 $\pm$ 0.011 & 0.385 $\pm$ 0.011 \\ \hline
    LRP: EpsilonGammaBox        & 0.339 $\pm$ 0.011             & 0.345 $\pm$ 0.011 & 0.340 $\pm$ 0.011 & 0.339 $\pm$ 0.011 \\ \hline
    LRP: EpsilonAlpha2Beta1Flat & 0.391 $\pm$ 0.011             & 0.391 $\pm$ 0.011 & 0.347 $\pm$ 0.011 & 0.347 $\pm$ 0.011 \\ \hline
    DeepLIFT                   & 0.366 $\pm$ 0.012             & 0.347 $\pm$ 0.010 & 0.354 $\pm$ 0.011 & 0.352 $\pm$ 0.011 \\ \hline 
    Guided Grad-CAM  & 0.278 $\pm$ 0.010     & 0.314 $\pm$ 0.011     & 0.351 $\pm$ 0.011     & 0.351 $\pm$ 0.011     \\ \hline                           
    LIME  & \multicolumn{2}{|c|}{\textbf{0.821} $\pm$ 0.010} & \multicolumn{2}{|c|}{\textbf{0.557} $\pm$ 0.014} \\ \hline 
    SHAP & \multicolumn{2}{|c|}{0.805 $\pm$ 0.010} & \multicolumn{2}{|c|}{\textbf{0.556} $\pm$ 0.015} \\ \hline
    Occlusion        & \multicolumn{2}{|c|}{0.367 $\pm$ 0.012}  & \multicolumn{2}{|c|}{0.391 $\pm$ 0.013}     \\ \hline \hline
    mean                        & 0.696 $\pm$ 0.010             & 0.677 $\pm$ 0.010 & 0.495 $\pm$ 0.013 & 0.496 $\pm$ 0.013 \\ \hline
  \end{tabular}
  \caption{Skinned: IROF $\uparrow$ - different channel poolings. Uncertainties are the standard error of the mean (383 images).}
  \label{tab:IROF_Skinned}
\end{table}

\begin{table}[htp]
  \centering
  \begin{tabular}{|l||l|l|l|l|} \hline
                        & Average Sensitivity $\downarrow$  & Complexity $\downarrow$ \\ \hline \hline
    Gradients        & 2.850 $\pm$ 0.035     & 9.143 $\pm$ 0.010     \\ \hline             Input x Gradients   & 1.807 $\pm$ 0.019     & 9.057 $\pm$ 0.010     \\ \hline          
    Integrated Gradients      & \textbf{1.407} $\pm$ 0.007     & 9.104 $\pm$ 0.011     \\ \hline   
    Deconvolution    & 2.170 $\pm$ 0.012     & 9.192 $\pm$ 0.011     \\ \hline             Guided Backpropagation   & 4.344 $\pm$ 0.081     & 8.746 $\pm$ 0.011     \\ \hline     
    LRP: EpsilonPlusFlat   & 1.756 $\pm$ 0.044     & 9.075 $\pm$ 0.014     \\ \hline       LRP: EpsilonGammaBox   & 2.001 $\pm$ 0.049     & 9.031 $\pm$ 0.012     \\ \hline       
    LRP: EpsilonAlpha2Beta1Flat    & 2.318 $\pm$ 0.066     & 8.902 $\pm$ 0.017     \\ \hline
    DeepLIFT         & 1.804 $\pm$ 0.014     & 9.116 $\pm$ 0.010     \\ \hline             Guided Grad-CAM    & 22.421 $\pm$ 7.328    & \textbf{8.640} $\pm$ 0.024     \\ \hline 
      \end{tabular}
  \caption{Skinned: Evaluation of quality of the explanations. Uncertainties are the standard error of the mean (383 images).}
  \label{tab:quality_skinned}
\end{table}

\begin{table}[htp]
  \centering
  \begin{tabular}{|l||l|l|l|l|} \hline
                                & mean pooling      & max pooling       & max abs pooling   & $\ell_2$-norm pooling \\ \hline \hline
    Gradients                   & 0.707 $\pm$ 0.002 & 0.838 $\pm$ 0.005 & 0.897 $\pm$ 0.005 & 0.897 $\pm$ 0.005     \\ \hline 
    Input x Gradients           & 0.709 $\pm$ 0.002 & 0.842 $\pm$ 0.004 & 0.903 $\pm$ 0.005 & 0.905 $\pm$ 0.005     \\ \hline 
    Integrated Gradients        & 0.709 $\pm$ 0.002 & 0.842 $\pm$ 0.004 & 0.903 $\pm$ 0.005 & 0.903 $\pm$ 0.005     \\ \hline
    Guided Backpropagation             & 0.707 $\pm$ 0.002 & 0.817 $\pm$ 0.003 & 0.890 $\pm$ 0.004 & 0.890 $\pm$ 0.004     \\ \hline 
    Deconvolution               & 0.685 $\pm$ 0.002 & 0.776 $\pm$ 0.003 & 0.809 $\pm$ 0.003 & 0.811 $\pm$ 0.003     \\ \hline
    LRP: EpsilonPlusFlat        & \textbf{0.908} $\pm$ 0.005 & \textbf{0.908} $\pm$ 0.005 & \textbf{0.917} $\pm$ 0.005 & \textbf{0.917} $\pm$ 0.005     \\ \hline
    LRP: EpsilonGammaBox        & 0.702 $\pm$ 0.003 & 0.765 $\pm$ 0.003 & 0.910 $\pm$ 0.005 & 0.910 $\pm$ 0.005     \\ \hline
    LRP: EpsilonAlpha2Beta1Flat & 0.768 $\pm$ 0.004 & 0.768 $\pm$ 0.004 & 0.906 $\pm$ 0.005 & 0.906 $\pm$ 0.005     \\ \hline
    DeepLIFT                    & 0.714 $\pm$ 0.002 & 0.839 $\pm$ 0.004 & 0.904 $\pm$ 0.005 & 0.904 $\pm$ 0.005     \\ \hline 
    Guided Grad-CAM  & 0.709 $\pm$ 0.003     & 0.816 $\pm$ 0.003     & 0.889 $\pm$ 0.005     & 0.889 $\pm$ 0.005     \\ \hline  
    LIME & \multicolumn{2}{|c|}{0.809 $\pm$ 0.008} & \multicolumn{2}{|c|}{0.887 $\pm$ 0.005} \\ \hline 
    SHAP & \multicolumn{2}{|c|}{0.738 $\pm$ 0.010}  & \multicolumn{2}{|c|}{0.772 $\pm$ 0.011} \\ \hline 
    Occlusion  & \multicolumn{2}{|c|}{0.754 $\pm$ 0.005}  & \multicolumn{2}{|c|}{0.905 $\pm$ 0.005}     \\ \hline  \hline
    mean                        & 0.758 $\pm$ 0.003 & 0.856 $\pm$ 0.004 & 0.910 $\pm$ 0.005 & 0.910 $\pm$ 0.005     \\ \hline 
  \end{tabular}
  \caption{Skinned: ROC-AUC $\uparrow$ for each explainability method and channel pooling type. Uncertainties are the standard error of the mean (145 images).}
  \label{tab:ROC_Skinned}
\end{table}

\begin{table}[htp]
  \centering
  \begin{tabular}{|l||l|l|} \hline
                                & max abs pooling   & $\ell_2$-norm pooling \\ \hline \hline
    Gradients        & 0.140 $\pm$ 0.005     & 0.138 $\pm$ 0.005     \\ \hline 
    Input x Gradients   & 0.350 $\pm$ 0.017     & 0.349 $\pm$ 0.017     \\ \hline 
    Integrated Gradients   & 0.346 $\pm$ 0.017     & 0.345 $\pm$ 0.016     \\ \hline  
    Deconvolution    & 0.055 $\pm$ 0.003     & 0.056 $\pm$ 0.003     \\ \hline
    Guided Backpropagation    & 0.246 $\pm$ 0.007     & 0.246 $\pm$ 0.007     \\ \hline    
    LRP: EpsilonPlusFlat   & \textbf{0.363} $\pm$ 0.010     & \textbf{0.363} $\pm$ 0.010     \\ \hline    
    LRP: EpsilonGammaBox   & 0.251 $\pm$ 0.007     & 0.249 $\pm$ 0.007     \\ \hline 
    LRP: EpsilonAlpha2Beta1Flat    & 0.336 $\pm$ 0.010     & 0.336 $\pm$ 0.010     \\ \hline                                  
    DeepLIFT         & 0.348 $\pm$ 0.016     & 0.347 $\pm$ 0.016     \\ \hline             Guided Grad-CAM    & 0.258 $\pm$ 0.010     & 0.257 $\pm$ 0.009     \\ \hline \hline 
    mean     & 0.338 $\pm$ 0.016     & 0.338 $\pm$ 0.016     \\ \hline                                                   
  \end{tabular}
  \caption{Skinned: Relvance Mass Accuracy $\uparrow$ for each explainability method and channel pooling type. Uncertainties are the standard error of the mean (145 images). Only max abs and $\ell_2$-norm pooling are included since this method expects non-negative explanations.}
  \label{tab:RelevanceMassAccuracy_Skinned}
\end{table}

\end{appendix}

\end{document}